%% file: main.tex
\documentclass{article}

\newif\ifdraft
\drafttrue % or \draftfalse
\draftfalse

\input{preamble}

\ifluatex
\usepackage[backend=biber,style=numeric,citestyle=numeric]{biblatex}
\fi

\usepackage{fullpage}

\title{Development of a New Image-to-text Conversion System for Pashto,\\
  Farsi  and Traditional Chinese\\
  \vskip1em
  \large{Machine Learning and Big Data Approach}}

\author{Marek Rychlik\footnote{University of Arizona, Department of Mathematics},
  Dwight Nwaigwe\footnote{University of Arizona, Department of Mathematics},
  Yan Han\footnote{University of Arizona, Library Sciences},
  Dylan Murphy\footnote{University of Arizona, School of Information}
}

\date{April 28, 2020}

\begin{document}
\maketitle
\begin{abstract}
  We report upon the results of a research and prototype building
  project \emph{Worldly~OCR} dedicated to developing new, more
  accurate image-to-text conversion software for several languages and
  writing systems. These include the cursive scripts Farsi and Pashto,
  and Latin cursive scripts.  We also describe approaches geared
  towards Traditional Chinese, which is non-cursive, but features an
  extremely large character set of 65,000 characters.  Our methodology
  is based on Machine Learning, especially Deep Learning, and Data
  Science, and is directed towards vast quantities of original
  documents, exceeding a billion pages.  The target audience of this
  paper is a general audience with interest in Digital Humanities or
  in retrieval of accurate full-text and metadata from digital images.
\end{abstract}

\tableofcontents

\marek{Please use this annotation mechanism.}

%\dylan{TEST: This is Marek on behalf of Dylan.}
%\yan{ }

\include{background}
\include{results}
\include{methods}
\include{software-overview}

% ----------------------------------------------------------------
\include{yan}
% ----------------------------------------------------------------

\appendix
\include{video-catalogue}

\ifluatex
\printbibliography
\else
\bibliographystyle{plain}
\bibliography{bibliography.bib}
\fi

\end{document}

%% file: preamble.tex
\usepackage{ifpdf}
\usepackage{ifpdf}
\usepackage{ifxetex}
\usepackage{ifluatex}

\usepackage{xspace}
\newcommand{\marek}[1]{}
\newcommand{\dylan}[1]{}
\newcommand{\dwight}[1]{}
\newcommand{\yan}[1]{}
\newcommand{\raymundo}[1]{}
\newcommand{\sindy}[1]{}

\ifdraft
\renewcommand{\marek}[1]{{\color{red} Marek says: #1\xspace}}
\renewcommand{\dylan}[1]{{\color{Violet} Dylan says: #1\xspace}}
\renewcommand{\dwight}[1]{{\color{Maroon} Dwight says: #1\xspace}}
\renewcommand{\yan}[1]{{\color{cyan} Yan says: #1\xspace}}
\renewcommand{\raymundo}[1]{{\color{NavyBlue} Raymundo says: #1\xspace}}
\renewcommand{\sindy}[1]{{\color{OrangeRed} Sindy says: #1\xspace}}
\fi

\usepackage{authblk}
\ifluatex

\usepackage[T1]{fontenc}
\usepackage{fontspec}

\newfontfamily{\arabicfont}{DejaVuSans}

\newfontfamily{\bromellofont}[ 
Extension = .ttf,
UprightFont = *-Regular]{Bromello}

\newfontfamily{\lunafreyafont}[ 
Extension = .ttf,
Scale=2]{Lunafreya}

\newfontfamily{\hanaminafont}[
Extension=.ttf,
Scale=1.2
]{HanaMinA}

\newfontfamily{\droidfont}[
Extension=.ttf,
Scale=1.2
]{DroidSansFallback}

\newcommand{\chinesefont}[1]{\droidfont{#1}}

\newcommand{\arabicfontorimg}[2]{\arabicfont{#1}}
\newcommand{\bromellofontorimg}[2]{\bromellofont{#1}}
\newcommand{\lunafreyafontorimg}[2]{\lunafreyafont{#1}}

\else 
\usepackage[T1]{fontenc}
\usepackage{CJKutf8}

\newcommand{\chinesefont}[1]{\begin{CJK*}{UTF8}{gbsn}#1\end{CJK*}}

\newcommand{\arabicfontorimg}[2]{\includegraphics[scale=0.2]{#2}}
\newcommand{\bromellofontorimg}[2]{\includegraphics[scale=0.3]{#2}}
\newcommand{\lunafreyafontorimg}[2]{\includegraphics[scale=0.3]{#2}}

\fi

\usepackage[dvipsnames]{xcolor}

\definecolor{Darkgreen}{rgb}{0,0.4,0}
\definecolor{Darkred}{rgb}{0.8,0,0}
\definecolor{Darkblue}{rgb}{0,0,0.4}

\usepackage[framemethod=TikZ]{mdframed}

\usepackage{pdfpages}
\usepackage{setspace}

\usepackage{fullpage}
\usepackage{multirow,tabularx,booktabs}
\usepackage{graphicx}
\usepackage{verbatim}
\usepackage{fancyvrb}
\usepackage{colortbl}
\usepackage[export]{adjustbox}% http://ctan.org/pkg/adjustbox

\definecolor{Gray}{gray}{0.9}

\usepackage[outline]{contour}% http://ctan.org/pkg/contour
\contourlength{0.1pt}
\contournumber{10}

\usepackage{subcaption}
\usepackage{caption}

\usepackage{wrapfig}            %For wrapping text around figures
\usepackage[vflt]{floatflt}
\usepackage{tikz}
\usepackage{tikz-3dplot}        %For 3d coordinate system and plots
\usepackage{tikz-qtree}
\usepackage{float} 

\usetikzlibrary{
  arrows,decorations.markings,
  positioning,shapes,
  automata,
  backgrounds,
  petri,
  shadows,
  arrows.meta,
  calc,
  plotmarks,
  matrix,
  shapes.geometric,
  trees
}

\usepackage{etoolbox}
\usepackage{amssymb}            %Math fonts are needed
\usepackage{relsize}            %For relatively sized fonts

\usepackage[inline]{enumitem}

\setlist{noitemsep}
\setlist[enumerate,1]{label=\textbf{(\arabic*)}}
\setlist[itemize,1]{label=$\bullet$}
\setlist[itemize,2]{label=$\diamond$}

\usepackage{comment}

\usepackage{hyperref}
\hypersetup{%
  pdfborder = {0 0 0},
  colorlinks,
  citecolor=black,
  filecolor=Darkgreen,
  linkcolor=black,
  urlcolor=red!40!black!70,
  linkbordercolor=red,      % hyperlink borders will be red
  pdfborderstyle={/S/U/W 1} % border style will be underline of width 1pt
}

%% file: background.tex
\section*{Acknowledgment}
This article has been made possible in part by the National Endowment for the Humanities: 
\emph{Development of Image-to-text Conversion for Pashto and Traditional Chinese}.

Any views, findings, conclusions, or recommendations expressed in this
article do not necessarily represent those of the National Endowment
for the Humanities.

\section{Background}
\subsection{Prologue}
The world's documents use many writing scripts and mainly consist of
Latin (4,900+ million pages), Chinese (1,340 million pages),
Arabic(660+ million pages), Devanagari (600+ million pages), Cyrillic
(250 million pages), Bengali (220 million pages), Kana (120 million
pages) and others. Latin languages use an alphabet, which is a
standard set of letters to form words. In contrast, the Chinese
language is a logographic system, of which a single written character
represents a complete grammatical word.

While developing OCR technology for printed Latin scripts is viewed as
a solved problem, as character and word accuracies typically reach
over 95\%, the problem for other languages, including Traditional
Chinese and Arabic scripts, is not yet completely solved. See
section~\ref{sec:literature-review} for a detailed discussion.  In
regard to Traditional Chinese, for documents utilizing modern fonts
OCR is generally a solved problem, but old Traditional Chinese
documents are still a big obstacle for OCR technology. An East Asian
studies librarian wrote to the author in 2019:

\begin{quote}
  I am just back from the annual AAS (Association for Asian
  Studies) and CEAL (Council on East Asian Libraries) meetings. This
  year (2019), Prof. Peter Bol of Harvard hosted a 2-day digital tech
  expo there to promote digital humanities... I spent 1 day on the DH
  sessions, where scholars constantly mentioned Chinese OCR as a
  conspicuous and serious block on their path to assessing ``digitized''
  textual collections. If you and your team succeed, it will surely help
  the EAS scholarly community a lot.
\end{quote}

\subsection{Our main objective --- building an Open Source OCR system}
The primary objective of this project is creation of a new, modern OCR
system, which will serve users with various needs. This includes the
individual users, such as a researcher who would like to perform OCR
on a single page or book, and insitutions who may have large
collections with millions of pages, such as a university or a library.
In order to satisfy such a diverse community, the software will be produced
in several forms:
\begin{enumerate}
\item A standalone \textbf{desktop software application} with a
  Graphical User Interface, running on \textbf{all major operating
    systems}, including Windows, Mac OS, Linux.
\item A \textbf{Web application} accessible through a Web browser and
  running on a server; the power of the server will determine the
  processing speed and quantity of data that can be handled.
\item A \textbf{MATLAB application}, primary for programmers and
  scientists who participate in the development of the system.
\end{enumerate}
The standalone application is ideal for individual users, with a
relatively small quantity of data. Large projects may use either a Web
application or the standalone application. Parallel processing will be
supported in all forms, which allows for shortening of processing time
by the use of higher end hardware (clusters, supercomputers, cloud).

This level of cross-platform support may seem like a tall order for a
small, research oriented team.  Therefore, it is worth explaining how
this is accomplished. We leverage the features provided by the MATLAB
commercial software \cite{matlab}, which allows us to deliver our
application in all of the above forms without extra work. Furthermore,
our prototype software \textbf{has been verified to work on those
  multiple platforms}.  In section~\ref{sec:matlab-platform} we
provide further details about how MATLAB is used in the project, and
how it will be used as a software delivery platform.

It should be noted that \textbf{the use of MATLAB does not make our
  software commercial} \footnote{Some colleagues have been concerned
  about this aspect of our software.}.  The relationship of our
software to MATLAB is similar to an Open Source application running
on a commercial operating system (Windows, MacOS). The availability of
MATLAB Runtime (the portion of MATLAB used by our standalone
application) and a clear license that allows free use, addresses this
matter
(\url{https://www.mathworks.com/products/compiler/matlab-runtime.html}).

\subsection{The definition of OCR and its advantages}
Image-to-text conversion, also called Optical Character Recognition
(OCR), is a crucial technology for Digital Humanities, connecting
historical documents written on traditional media (e.g., paper, cloth)
to modern, digitized storage. The main problem of the field sounds
deceptively simple: to take an image, e.g., a photograph of a page of
printed or handwritten page of text, and retrieve the original
text. More precisely, we essentially reproduce the \textbf{sequence of
  keystrokes} that would result in text \textbf{semantically
  equivalent} to the content of the original page of text. Of course,
a human expert familiar with the writing system used, and capable of
typing or other form of data entry, may perform this task. The
challenge is to \textbf{teach a machine} to perform this task and
scale up the speed and volume that can be handled, so that the
billions of pages of historically important written records can be
converted from image form to text.

The benefits of having the text are numerous. Foremost, the text can
be searched and edited with standard computer software, such as
Microsoft Word or Emacs (our favorite text editor). Furthermore, the
text can be processed with \emph{Natural Language Processing} (NLP) software
to study it from the linguistics point of view. 

There are additional benefits to society, such as reduction of disk space
used to store a \textbf{document's meaning} rather than various artifacts of its
processing, such as yellowing of pages, chipping of ink, warping of paper, etc.
Easily, the savings of disk space may be a \textbf{thousand-fold}, which may
be translated into similar saving of, say, power consumption.

In short, OCR facilitates access to and reduces the cost of storing of
massive quantities of data that is used in Digital Humanities, and
will be multiplied rapidly in the near future.

\begin{comment}

  \dwight{This section seems more appropriately placed close to the
    beginning of the paper, such as after the Image-to-Text Conversion
    section. That subsection is a good lead-in for this one.}

\marek{Indeed. It was moved}

\end{comment}
\subsection{Artificial Intelligence, Machine Learning and Data Science}
As hiring human experts is prohibitively expensive for the totality
of records under consideration, the conversion has to be performed
with software imbued with the expert knowledge required to perform
image-to-text conversion. Artificial Intelligence (AI) is defined
as constructing machines or software capable of performing tasks
at a level of a competent human. Therefore, image-to-text conversion
is considered a branch of AI, and makes the current project an AI project.

Recent rapid progress in AI is mainly due to advancements in many
distinct areas.  In particular, Machine Learning (ML) has become a
feasible approach to many AI problems. The premise of ML is to build
software by training rather than programming. The components of the
software that learn and retain information from the training are
called (Artificial) Neural Networks, and are inspired by the
biological neural network: the brain. Thus, a human programmer (e.g.,
a member of the current team), builds software which trains a neural
network to perform image-to-text conversion. Neural networks are
typically trained using \textbf{supervised learning}, whereby a
sequence of examples is presented to the network of a task at hand
performed correctly, and the network generalizes the examples to a
much larger body of problems. In image-to-text conversion the network
may be presented with a few pages of correctly transcribed text, and
the network is then capable of transcribing all similar texts.

\subsection{Recent successes with cursive scripts}
The significant recent progress in OCR of cursive scripts
(e.g., Arabic, Farsi, Pashto, off-line handwriting) has been mainly
achieved due to progress in Machine Learning (ML) and
especially Deep Learning (DL). All modern OCR systems incorporate ML
and DL components.

\begin{comment}
\marek{Yan, this is mostly unnecessary, and indeed would have
  to be polished to be included:
  
  ``Published articles propose multiple neural
  networks, including RNN, CNN, LSTM, BLSTM, CTC and etc. (citations)''
}
\end{comment}

Without going into the technical details associated with this approach
we hope to give the user a flavor of these techniques and explain the
impact on real workflows associated with them, as it relates to
practical image-to-text conversion. The particular aspect of ML is
that the software needs to be trained in addition to being
written. Training is a crucial phase of building a software system
which will perform well on real data.  The crucial part of this phase
is \textbf{collecting and digitizing samples of documents} and
\textbf{using human experts} to manually transcribe the documents to
Unicode (Unicode is described in
section~\ref{sec:overview-of-ocr-tech}), which makes it slow or
expensive, or both. Also, training may involve running software for
weeks to optimize hundreds of thousands of parameters, and any
interruption of this process, or mistakes in software design often
require starting from scratch. Hence, one of the important aspect of
this project is \textbf{shortening the length of time required to
  train} by using better algorithms and training protocols.

\subsection{An overview of the OCR technology}
\label{sec:overview-of-ocr-tech}
There are two major phases of converting a physical record, such as a
book printed hundreds of years ago, to a full-text digital record,
without actually having access to the physical record, but requiring the
record's textural content:
\begin{enumerate}
\item Digital capture, such as scanning with a digital scanner
  or digital camera, and storing the results in the form of
  a digital image.
\item Extraction of textual information from the image and storing it
  as encoded text, typically Unicode , either
  along with the original image, or replacing the digital image.
\end{enumerate}
Unicode, in simple terms, is numbering of the keys of a virtual
computer keyboard capable of producing all characters used in all
languages, also called \textbf{encoding}. The endeavor of creating
such numbering requires a \textbf{standards body}, and the  \emph{Unicode
  Consortium} \cite{unicode-consortium} is such a body. It should be
mentioned that many (often incompatible) encoding standards have been
created in the past, starting with the American Standard Code for
Information Interexchange (ASCII) first published in 1963, and Unicode
is designed to be ``the last'' standard, to be universally used in the
foreseeable future.

Of the two steps to OCR, \textbf{digital capture has become easy} due to the
development of inexpensive, miniaturized digital cameras of extremely
high quality.

The difficult part is the second step: image-to-text conversion.
It is paramount that this step be fully automated, as it must be
applicable to the totality of documents representing the cultural
heritage of entire nations. In principle, image-to-text conversion
can be performed by a large team of human experts, who visually
examine and recognize the text, and key-in the results using
a computer keyboard. The degree of expertise required
varies. For instance, texts in Traditional Chinese may be hundreds
of years old, and may require linguistic and historical knowledge
to perform accurately.

OCR systems evolved towards a consensus architecture that identifies
several common phases. This happened over a period of more than 40 years
since the digital scanner was invented and the first algorithms for performing
OCR on Latin scripts were developed
\cite{kurzweiltech,abbyy-ocr-history}.  OCR phases may include: image
pre-processing; image segmentation; recognition proper;
post-processing; linguistic analysis. A good review of the
architecture of the Tesseract~OCR system \cite{tesseract} can be found
in an excellent blog article entitled ``Breaking down Tesseract OCR''
(\url{https://machinelearningmedium.com/2019/01/15/breaking-down-tesseract-ocr/}).
Many themes of the article can be traced back to an earlier article by
Ray~Smith \cite{smith_overview_2007}.  Tesseract~OCR development
started over 30 years ago as a commercial project at HP and it is
still being developed by its original developer Ray~Smith. Tesseract
and various publications based on it represent both current and
historical views of the art of OCR. Kraken is a \emph{fork} of an older
system, \emph{Ocropus} \cite{ocropus}, released in 2007.  The project
does not appear to be actively developed, based on GitHub activity
statistics. Therefore, there is a need for a new, agile platform for
research and software development, and we hope that our platform will
fulfill this role.

\subsection{Rationale for building a new OCR system}
Our project has as its goal building a superior OCR system, with
higher accuracy and capacity to process huge quantities of data
quickly. We aim to be able to achieve 90\% accuracy, which is
considered a threshold for the OCR results to be useful, on a much
larger class of documents than currently possible.  Increasing OCR
accuracy is a particularly ambitious goal, as it involves advancing a
field developed over a period of 40 years by extremely strong
researchers. The strength of our team is its diverse yet complementary background in
mathematics, statistics and scientific computing.  The current focus
of the project is \textbf{fundamental research and software
  prototyping}.  The focus will gradually shift towards
\textbf{software development}, while maintaining the ability to
quickly prototype and do OCR research. We needed a software platform
which allows this approach. The natural choice is MATLAB
\cite{matlab}.  The pros and cons (mostly pros) of MATLAB as a
development platform are further discussed in
section~\ref{sec:matlab-platform}.

We did look at other ways to structure our project. We quickly
excluded commercial systems as they are essentially closed to
modifications by non-commercial developers.  Two open source domain
projects \emph{Tesseract} \cite{tesseract} and \emph{Kraken}
\cite{kraken} are open to other developers, but they are primarily
software development projects, not research platforms. This would
hinder our ability to have a comprehensive look at the totality of OCR
algorithms and bring in research results to bear on the problem.

\subsection{A review of prior OCR related research}
\label{sec:literature-review}
In the current section we briefly review mostly recent works devoted
to the core OCR problems, and thus closely related to our own
research, and software prototype development. Most of the problems
discussed have already been solved in our software
prototype. \textbf{We have not based our implementation on any
  particular work}, but designed our own algorithms. Naturally, our
methods overlap with much other research in the field. Often, rather
than working from scratch, we build our algorithms on top of the
MATLAB framework, which provides a vast collection of high quality
building blocks for a project such as ours.

\begin{comment}

\dwight{I don't think the paragraph below is that important. It seems
  too wordy. I suggest that instead of listing 2 competing process
  descriptions, we pick one. Based on my understanding of the original
  paragraph, I would rephrase it as:
}

\marek{We went with Dwight's suggestions. In addition, the list of
  phases of processing is no longer relevant or complete, so it is
  better to stick to generalities here.}
\end{comment}

\begin{mdframed}[
  middlelinecolor=blue!20,
  middlelinewidth=2pt,
  backgroundcolor=green!10,
  roundcorner=5pt]
   {\it The reader may skip the remainder of this section upon first
     reading without loss of understanding.}
\end{mdframed}

Tomaschek \cite{tomaschek_evaluation_nodate} identifies the following
stages of processing in an OCR system: image acquisition,
preprocessing, binarization, page segmentation, line, word, and
character segmentation, recognition, and post-processing. In contrast,
the dissertation \cite{shafii_optical_2014} describes the process as
having two major phases: preprocessing and recognition in a linear
sequence. The preprocessing phase consists of noise removal,
binarization, skew detection, page analysis, and segmentation, while
recognition phase includes feature extraction and classification. The
author suggests that context analysis and recognition are responsible
for final output, i.e., an advanced OCR system shall be more than just
text labeling.

\begin{comment}
  Over the past 40 years the architecture of OCR systems evolves towards
  a consensus architecture that identifies several common
  phases. \cite{tomaschek_evaluation_nodate} summarized stages: image acquisition,
  preprocessing, binarization, page segmentation, line, word, and
  character segmentation, recognition, and post-processing. In contrast,
  the dissertation \cite{shafii_optical_2014} describes the process as having two
  major phases: preprocessing and recognition in a linear sequence. The
  preprocessing phase consists of noise removal, binarization, skew
  detection, page analysis, and segmentation, while recognition phase
  includes feature extraction and classification. The author suggests
  the following phases where context analysis and recognition are two
  branches which are responsible for final output. An advance OCR shall
  be more than just classification (text labeling). With current
  advancement in AI and computer vision, semantic information along with
  character recognition provides more contextual information.

  \begin{enumerate}
  \item Image preprocessing,
  \item Segmentation,
  \item Post-processing,
  \item context analysis and Recognition.
  \end{enumerate}
\end{comment}

In recent years new systems based on Deep Learning ideas merged
various phases to reflect this new change in architecture.  In the
past various algorithms were proposed generally following the above
design. These are relevant to this day, and they can be combined with
the Deep Learning approach in various ways, and therefore are worth
discussing. The new development in OCR technology is an application of
Recurrent Neural Networks (RNN) which led to breakthroughs in OCR of
cursive scripts, starting with Arabic.  In the past, OCR systems
performed poorly over complex scripts such as Indian and Arabic
languages, and with RNNs the situation is greatly improved. The
successes of RNN in this area can be traced back to the seminal
research described in papers of Sepp~Hochreiter, J\"urgen~Schmidhuber
and Alex Graves \cite{lstm,graves_novel_2009,graves_offline_2008}. The
software system \emph{Kraken} \cite{kraken} referenced in other parts
of this paper is perhaps the most pure implementation of this circle
of ideas.

\begin{comment}
\dwight{I think we should state the scripts that each technology is suited for}

\marek{There are general purpose techniques, so no need to go into details}
\end{comment}

In the preprocessing stage, background and foreground segmentation
occurs using mostly statistical modeling.  The segmentation involves
\textbf{blobs}, also called ``objects'', which conceptually represent
the contiguous regions of a page covered by ink.  Using various
heuristics, one can combine blobs into larger entities: \textbf{lines
  of text}, \textbf{words}, \textbf{ligatures} and
\textbf{characters}, without detailed knowledge about the the
underlying languages, just on the basis of common principles used by
most writing systems. One major distinction is between scripts which
use separated characters (e.g., printed Latin scripts and the Chinese
logographic system), and cursive scripts, in which characters are
connected (e.g. handwritten Latin scripts, Pashto and Persian/Farsi).

Ciresan et. al. \cite{ciresan_convolutional_2011} proposed using a
statistical \emph{Gaussian mixture model}, and investigated simple
training data preprocessing and focused on improving recognition
rates using \textbf{committees of neural networks}. They suggested
using committee-based classifiers as basic building blocks of any OCR
system.

Dutta \cite{dutta_robust_2012} proposed recognizing character
$n$-gram images, which are groupings of consecutive
character/component segments. They use the character $n$-grams as a
primitive for recognition rather than for post-processing, resulting
in a 15\% decrease in word error rate on heavily degraded Indian
language document images.

Shafii \cite{shafii_optical_2014} presented methods for detecting skew
angle and reviewed multiple methods proposed before. One method is
based on geometrical features of a skewed document. The proposed
algorithm and its implementation show 95\% success rate on a certain
set of documents.

Li, Zheng, and Doemann \cite{yi_li_detecting_2006} proposed the use of a
\emph{Gaussian window} to detect text lines in handwritten documents.

Louloudis et.al \cite{louloudis2006block} presented a 3-step method
for text line detection. The first step includes image binarization
and enhancement, connected component (blob) extraction, partitioning of the
connected component domain into 3 spatial sub-domains and average
character height estimation. The second step is to use Hough transform
for the detection of potential text lines. The third step is to
correct possible splitting and to detect text lines not found by the
second step, and finally to separate vertically connected characters
and assign them to text lines.

The dissertation \cite{shafii_optical_2014} contains a literature
review related to page segmentation, including top-down algorithms,
bottom-up algorithm, and a combination of both.  Shafii proposed a new
segmentation technique based on image resampling methods. The method
simply performs a downsampling (zoom-out) by upsampling with a
calculated scaling factor. The sampling scale is calculated by
utilizing a white gap transition algorithm and determining between the
text row gaps. This process converts the image to blobs of
segments. Each segment is identified as text or non-text using three
criteria. Shafii concluded that many algorithms deemed to be successful
for English do not apply to Persian (83,84). Citing previous research on this topic, Shafii proposed a hybrid feature
extraction algorithm and 1st Nearest Neighbor classification for
sub-word segmentation.

\begin{comment}
  \sindy{References in following need to be fixed}
  \marek{Fixed}
\end{comment}

In the phase of post-processing, most popular approaches to handle
degraded documents are to use post-processing methods such as
character error models \cite{kahan_recognition_1987}, dictionaries \cite{lehal_shape_2001}, 
statistical language models \cite{natarajan_bbn_2010}, and/or a combination
as studied by Taghva and Sofsky \cite{taghva_ocrspell_2001}.

In the phase of recognition, Sankaran and Jawahar
\cite{sankaran_recognition_2012} proposed BLSTM (Bidirectional
Long-Short Term Memory) for the Indian script of Devanagari.

\begin{comment}
\dwight{maybe we should write CTC instead of RNN/Bi-LSTM,
  or combine both terms}

\marek{CTC may not have been used; needs checking}
\end{comment}

This approach does not require word to character segmentation, which
is one of the most common reasons for high word error rate. They
reported a 20\% reduction in word error rate and 9\% reduction in
character error rate when comparing with the best available OCR
system.

\begin{comment}
\sindy{"simple" meaning the BLSTM-CTC ?}
\marek{Addressed by adding a footnone}
\end{comment}

Paul and Chaudhuri \cite{paul_blstm_2019} presented a OCR system using
a single hidden BLSTM-CTC\footnote{BLSTM-CTC stands for
  ``Bidirectional Long-Short Term Memory'' and ``Connectionist
  Temporal Analysis''.}
architecture having 128 units. This architecture was trained by 47,720
text lines and tested over 20 different Bengali fonts, producing 99\%
character accuracy rate and 97\% word accuracy.  Using CNN-BLSTM-CTC
architecture, the results show the superiority of this architecture
over the simple architecture.  They reported an 89\% character
accuracy on the degraded image simple.

The International Conference on Document Analysis and Recognition
hosted a Chinese (simplified) handwriting competition in 2013 to gauge
the effectiveness of various technologies. Almost all entrants used
quadratic discriminant analysis or convolutional neural networks (CNN). The
most accurate systems in terms of isolated character recognition were
CNN-based, with accuracies of around 94\%. This was closely
followed by quadratic discriminant function-based methods which
obtained accuracies of around 92\% \cite{yin_icdar_2013}.

\subsection{MATLAB as a development platform}
\label{sec:matlab-platform}
There is considerable interest in applying machine learning to a variety of
problems of great importance to society. One well-known application is
building a self-driving car. Another area is applications to medicine,
in which a computer may look at images of lesions or tumors, and
derive a diagnosis, which is now often more accurate than that of a
human expert.  Image-to-text conversion shares a great deal of
technology and knowledge with these applications, as computer vision
is an essential ingredient. Indeed, our software looks at digital
images. Therefore, a significant effort has been devoted to providing
platforms on which to build software supporting all these applications.

\begin{comment}
  \sindy{The above last 2 sentence seemed like an unfinished thought. I changed. Check it.}
  \marek{Addressed by the last sentence}
\end{comment}

Our choice for implementing the \emph{Wordly OCR} prototype software
is MATLAB. This is commercial software with a generous academic license.
The software is well-known to engineers as an essential tool in
most engineering projects. MATLAB is also used extensively in the
Mathematics Department and across the sciences. Perhaps to lesser degree
it is a tool for humanities researchers, but is not unknown. Moreover,
due to its excellent documentation and commercial support, it is
easy to incorporate into humanities projects. Therefore MATLAB was
a natural choice for the current project. Moreover, MATLAB
features the Deep Learning Toolkit, which implements most of the
components needed in the current project.

\subsection{MATLAB as a software delivery system}
\textbf{One important aspect of MATLAB is the ability to deliver free
  software to the end user.} This is because the core of MATLAB code
is available as a collection of shared libraries called the
\emph{MATLAB Runtime}
(\url{https://www.mathworks.com/products/compiler/matlab-runtime.html})
and freely distributable with an application built with MATLAB. This
software distribution model is essentially the same as that of
Microsoft products built with Visual Studio.  Furthermore, 
\textbf{an application built with MATLAB is portable across all major
  OS platforms}. Also, it can be run as a \textbf{Web application},
utilizing the MATLAB server component. This means that a university or
an organization is able to provide on-line access to the application
running on a much more powerful computer than an average user can
afford (a supercomputer or a large cluster), e.g., to support
conversion of a large collection of documents quickly.  Thus, in
principle, MATLAB is an \textbf{ideal software delivery platform}.
Some limitations are due to the fact that this is a relatively new
addition to MATLAB.

\begin{figure}[htb]
  \centering
  \begin{subfigure}{0.85\textwidth}
    \includegraphics[width=\textwidth]{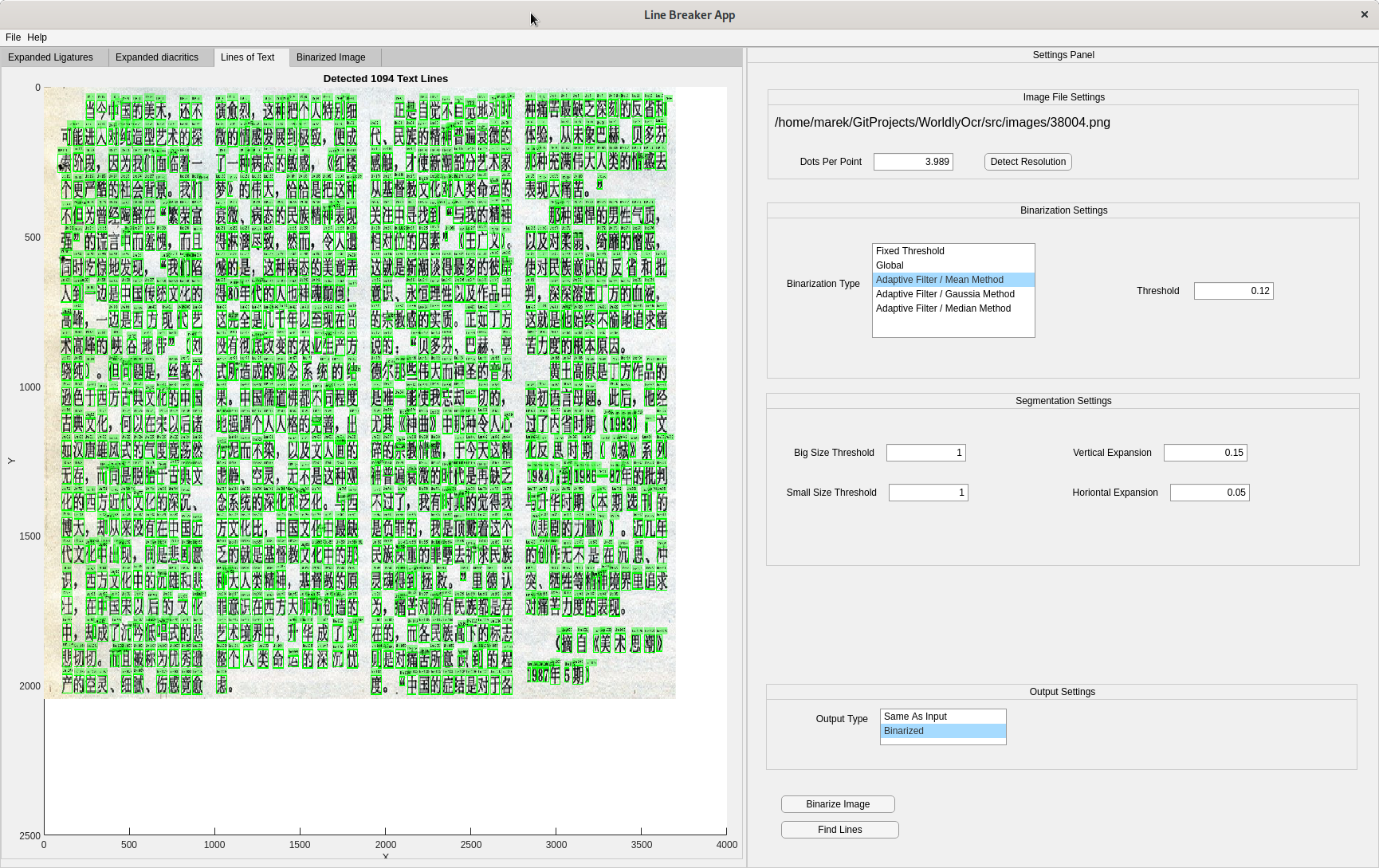}
    \caption{\label{fig:breakup-gui-chars} Basic view}
  \end{subfigure}

  \begin{subfigure}{0.85\textwidth}
    \includegraphics[width=\textwidth]{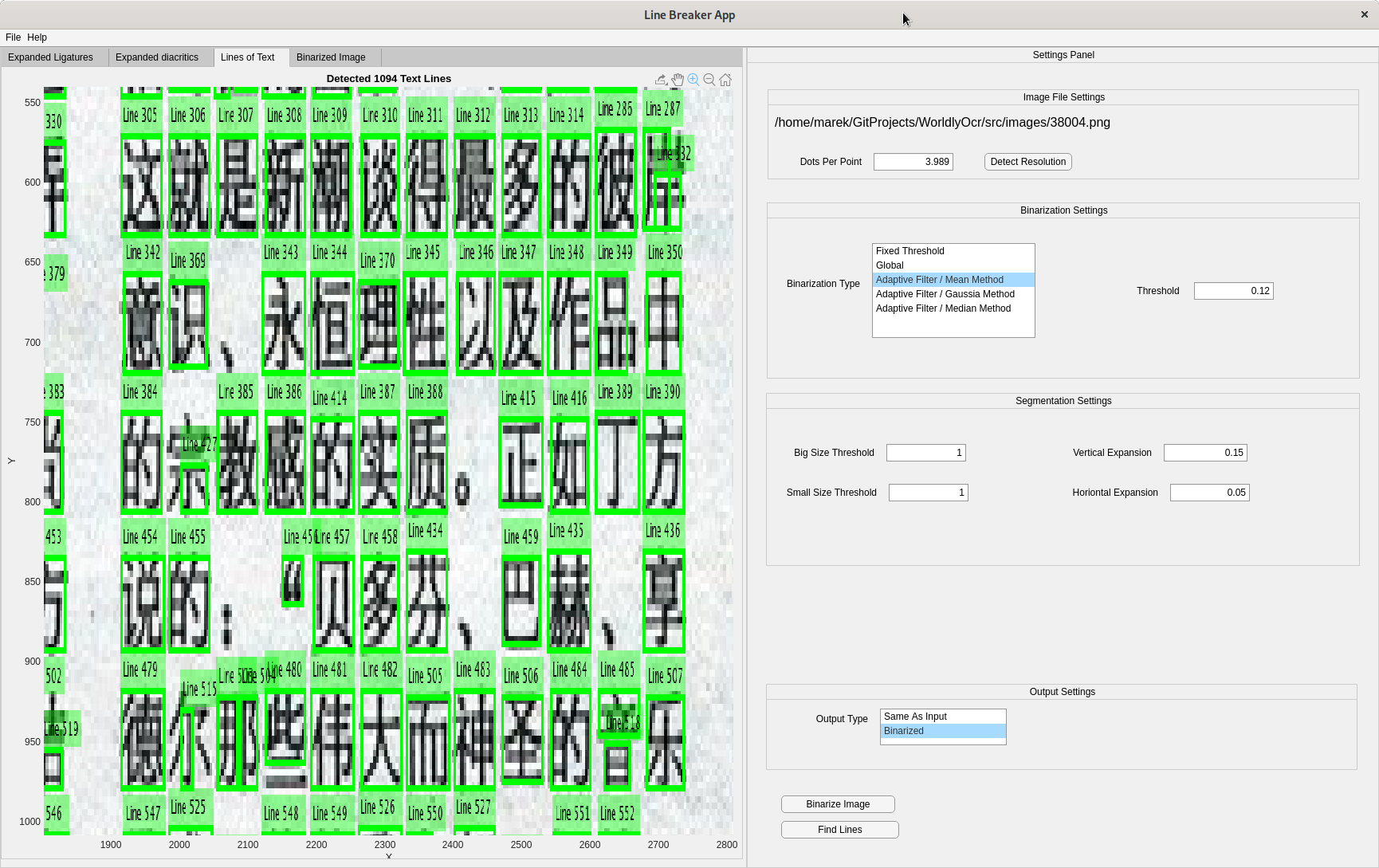}
    \caption{\label{fig:breakup-gui-blownup} Blown-up view}
  \end{subfigure}
  
  \caption{\label{fig:breakup-gui} A GUI-based program for breaking
    up pages into lines or characters. The GUI offers access to underlying
    parameters which can be interactively adjusted to parse new content
    and reused.}
\end{figure}

As an example, we developed an application for breaking up a page of
text into lines and characters. As indicated in other parts of the
paper (e.g., Figure~\ref{Fig-8}), complex layouts often lead to useless
OCR results, and therefore it makes sense to interactively aid the
software in determining the high-level document
organization. Therefore, we developed a GUI application which quickly
allows us to extract lines of text and characters from a page of text (an image).
The basic look of the GUI is shown in Figure~\ref{fig:breakup-gui}. So far, the
application has been used internally by the project to:
\begin{enumerate}
\item extract lines from Pashto documents to build a training dataset similar
  to the OCR\_GS\_Data dataset in Farsi;
\item extract characters from Chinese newsprint (Figure~\ref{fig:breakup-gui}).
\end{enumerate}
In the course of the project several GUI-based applications were constructed
for various tasks, and it is expected that some of them will become a part of
an integrated, GUI-based software package.

\subsection{Other platforms}
In the Data Science community there are many projects utilizing
Python, and ML libraries such as PyTorch and TensorFlow. A part of
their appeal is that they are Open Source, and are used to implement
many of the newest research ideas in the field. However, this comes at
a cost of lower stability, poorer documentation and a higher learning
curve. 

\subsection{Collaboration with our project}

\begin{comment}
\sindy{I changed the following sentence. Check it.}
\marek{Fine.}
\end{comment}

The potential user of our prototype software will find that it is relatively
easy to modify and add to the \emph{Worldly OCR} software with only
minimal knowledge of the underlying science. It is fair to say that
any scientist who has taken a course in Linear Algebra in which MATLAB
was used, is prepared to modify and adapt the code of \emph{Worldly
  OCR} for their projects.

%%% Local Variables:
%%% mode: latex
%%% TeX-master: "main.tex"
%%% TeX-engine: luatex
%%% End:

%% file: results.tex
\section{Results}
To date the results of the project are preparatory in nature to producing a full-fledged
software system. They generally fall into the following categories:
\begin{enumerate}
\item Fundamental OCR-related research, including original algorithms and examination
  of existing algorithms, for various phases of processing.
\item Prototyping parts of the system in MATLAB, ranging from small ``throw-away'' scripts,
  larger libraries of code, to complete applications with a Graphical User Interface (GUI).
\item Examination of existing and creation of new training datasets in
  Pashto, Farsi and Chinese.
\end{enumerate}
Generally each of the results discussed below contains parts which belong to all of these
categories.

\subsection{OCR on  1023 lines of text in Farsi with 97\% accuracy}

\begin{comment}
  \sindy{Is Pashto THE primary target or one of the primary targets?}
  \marek{I changed "primary target" to "next major target"}
\end{comment}

Farsi is a stepping stone towards processing Pashto, which is the
next major target for our software. There is no significant fundamental
difference between the two languages in regard to the writing
system. However, there is a logistical difference: the existence of
verified training data (``the gold standard'') for Farsi. As creating
a comparable standard for Pashto would be costly in time and
resources, our prototype software is trained on existing Farsi data,
to be later on applied to Pashto, when suitable training data becomes
available or is created.

We trained our OCR software to decode 1023 lines of Farsi text
achieving \textbf{approximately 97\% accuracy} with 2 hours of
training on a laptop computer. The input data was produced by using
MATLAB Unicode text rendering in a specific font: DejaVu Sans. This font
has good support for Arabic scripts and is typically present on Linux
systems, but can be downloaded from the DejaVu font project as a
True~Type font (.ttf) file \cite{project_dejavu_2020}.

\begin{figure}[tbh]
  \centering
  \includegraphics[width=0.8\textwidth]{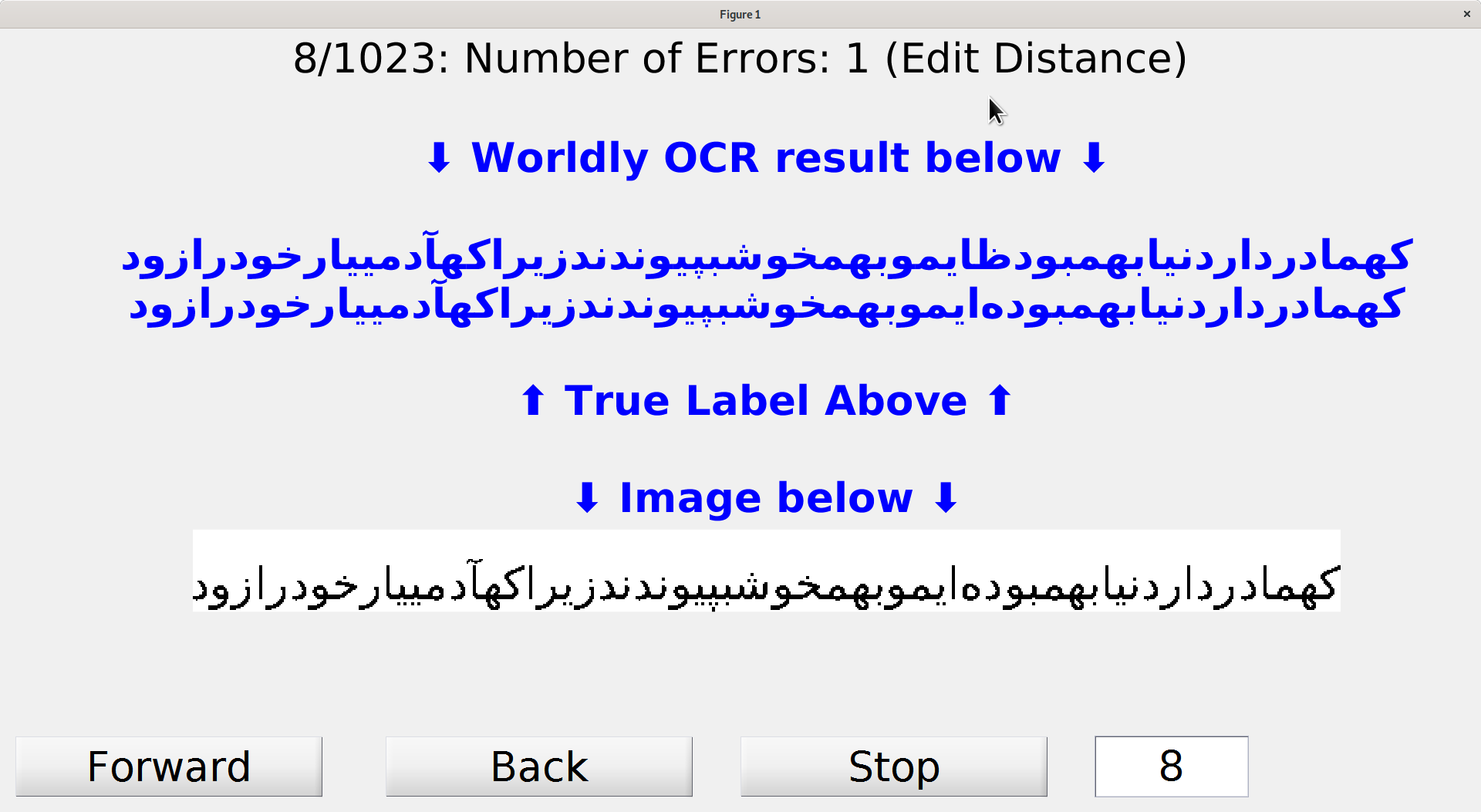}
  \caption{The OCR result obtained with our software, compared to the true label
    and the image input. This example features 1 error (1 omitted character).
    The number of errors is measured automatically using \emph{edit distance},
    i.e., the minimal number of single-character deletions, insertions and substitutions
    to change one string into another.
    \label{fig:farsi-typical-line}}
\end{figure}

The full results of our experiment have been presented in the form of
a video currently posted on YouTube \cite{rychlik-farsi-ocr-video}.
One example is shown in Figure~\ref{fig:farsi-typical-line}. It should
be noted that an average line of text has 60+ characters, and thus one character
error per line is roughly equivalent to a 1.6\% error rate.

\begin{figure}[bth]
  \centering
  \includegraphics[width=.9\textwidth]{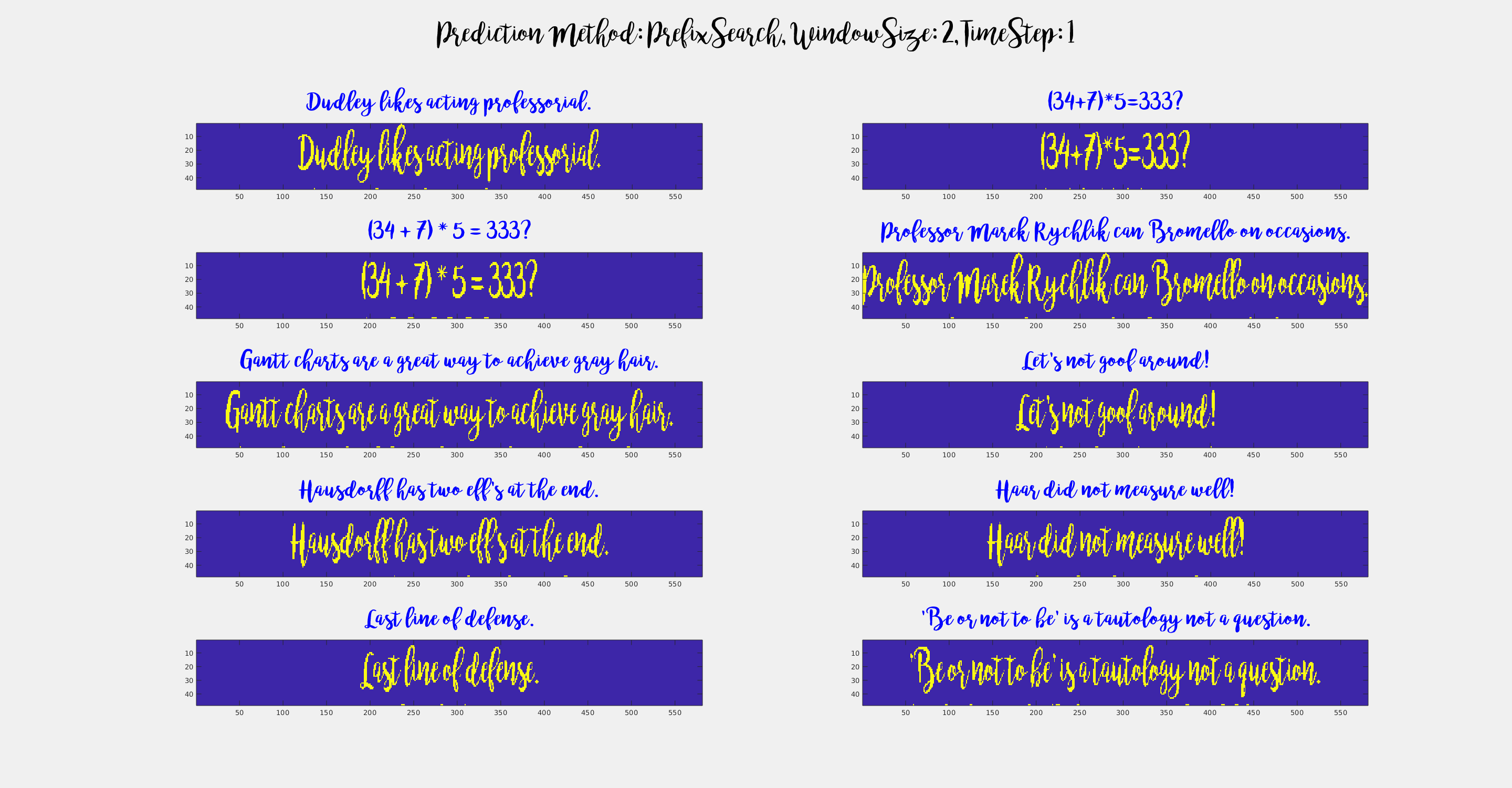}
  \caption{Our OCR results on samples utilizing the Bromello
    font. Accuracy is 100\% \textbf{including whitespace}. Labels
    obtained with our OCR are atop of OCR'ed images picturing
    yellow-on-blue text.\label{fig:bromello-sample}}
\end{figure}

\subsection{OCR on Latin cursive script Bromello}
In our research we used a Latin cursive script font, Bromello, as a
stepping stone to handling non-Latin scripts, besides being of
interest as an ML  research problem. One standard line of Bromello looks like this:
\begin{center}
  \begin{large}
    \bromellofontorimg{The quick brown fox jumped over the lazy dog.}{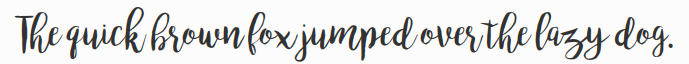}
  \end{large}
\end{center}
Bromello is not the only Latin script font. We also experimented with
a font called Lunafreya, which is slanted:
\begin{center}
  \begin{Huge}
    \lunafreyafontorimg{The quick brown fox jumped over the lazy dog.}{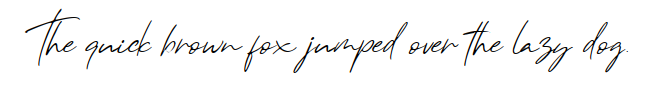}
  \end{Huge}
\end{center}
At this time we do not have results for Lunafreya to report.

Figure~\ref{fig:bromello-sample} shows several Bromello samples
subjected to our OCR software.  Latin cursive scripts are not used in
normal typesetting, but are used in specialty printing, such as
decorative documents (e.g. greeting cards). Latin script exhibits some
features of non-Latin scripts, but it does not present the full range
of difficulties, because it was designed for easy implementation along
traditional typesetting systems for Latin script.

\begin{figure*}[htb]
  \begin{subfigure}[t]{0.5\textwidth}
    \includegraphics[width=\textwidth]{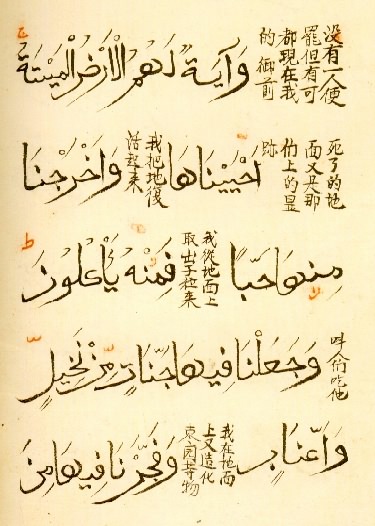}
    \caption{\label{fig:chinese-quran}Chinese Qur'an in Sini with Chinese translation}
  \end{subfigure}
  \begin{subfigure}[t]{0.5\textwidth}
    \includegraphics[width=\textwidth]{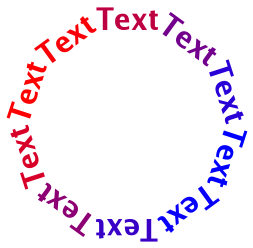}
    \caption{\label{fig:rotated-text}Random rotated Latin script found on Google}
  \end{subfigure}
  \caption{\label{fig:challenging-documents} Two types of text
    written neither horizontally nor vertically. Chinese Qur'an source:
    \url{https://en.wikipedia.org/wiki/Islamic_calligraphy}; this
    document breaks many rules of typical Arabic texts.}
\end{figure*}

Our software attains \textbf{100\% accuracy} on a great majority of
texts typeset utilizing Bromello. The study of Bromello and Lunafreya
Latin fonts is hoped to help with developing models of typed and
handwritten text which are required to convert documents such as the
one in Figure~\ref{fig:chinese-quran}.

\subsection{Cursive scripts}
One major theme of our project is image-to-text conversion of cursive
scripts.  The term OCR is used but is not quite correct in the context of
cursive scripts, as no character recognition in isolation is actually
performed, but groups of characters, typically lines of text, need to
be recognized. Nevertheless, for convenience we will use the
abbreviation OCR for the family of algorithms and softwares under
consideration.

Of particular interest are cursive scripts used throughout
Middle East and Western Asia, such as Arabic, Farsi and Pashto. This
paper describes building an actual implementation of the algorithms
under discussion as part of a project sponsored by the National
Endowment for Humanities (NEH). The software will be applied to a
large dataset \cite{afghandata} consisting of approximately 2 million
pages of documents in Pastho and Farsi (sometimes mixed in the same
document).  Altogether, these documents represent a substantial
portion of the cultural heritage of Afghanistan. The documents are a
product of long term collaboration between the University of Arizona
and Kabul University (ACKU), including digital scanning of the documents.
We will refer to this collection as ACKU dataset. The experience acquired
by working with these large datasets will allow us to develop strategies
to be applied to other languages and large-to-huge datasets.

\begin{figure*}[htb]
  \begin{subfigure}[t]{0.5\textwidth}
    \includegraphics[width=\textwidth]{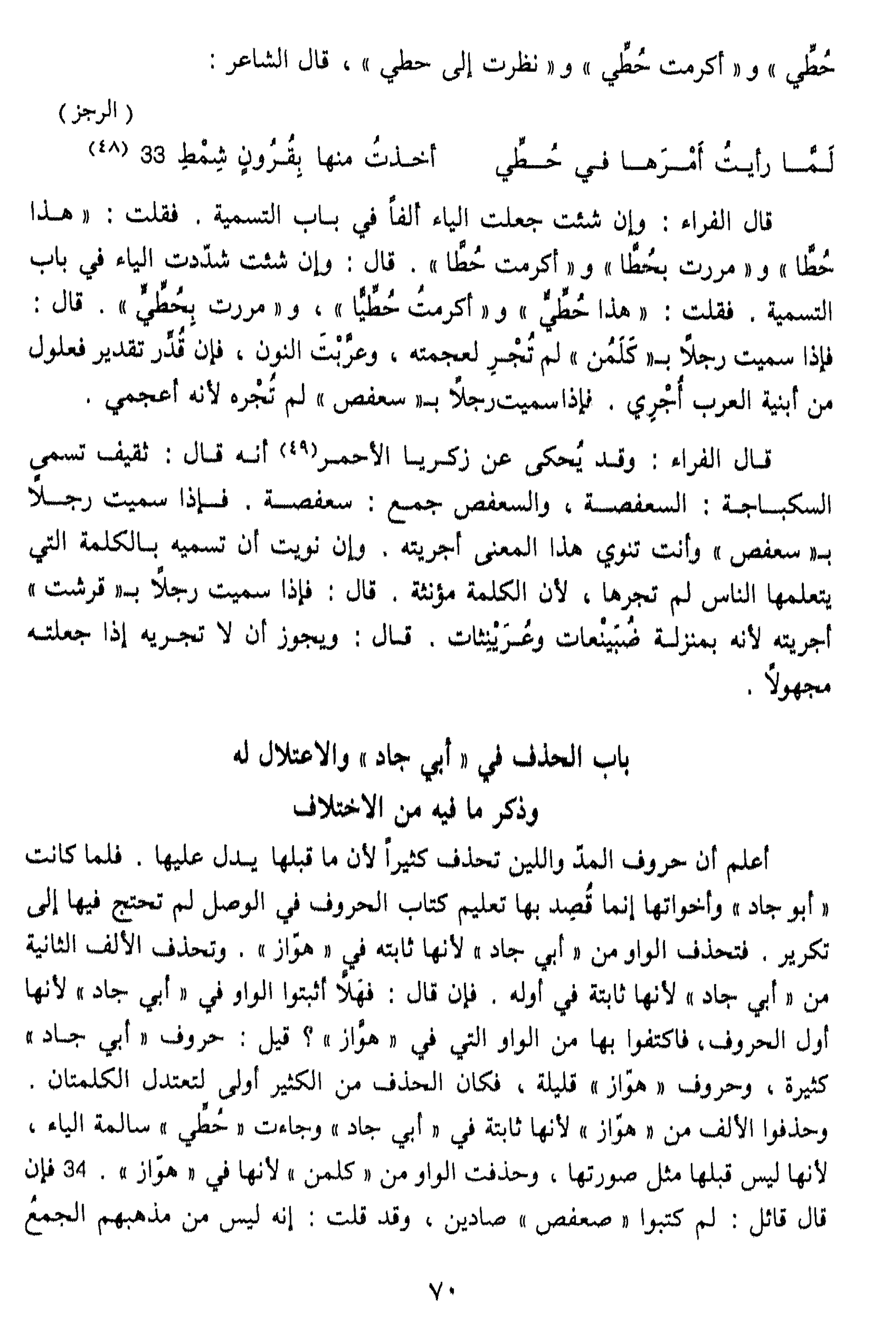}
    \caption{\label{fig:arabic-segmentation-input} A sample page}
  \end{subfigure}
  \begin{subfigure}[t]{0.5\textwidth}
    \includegraphics[width=\textwidth]{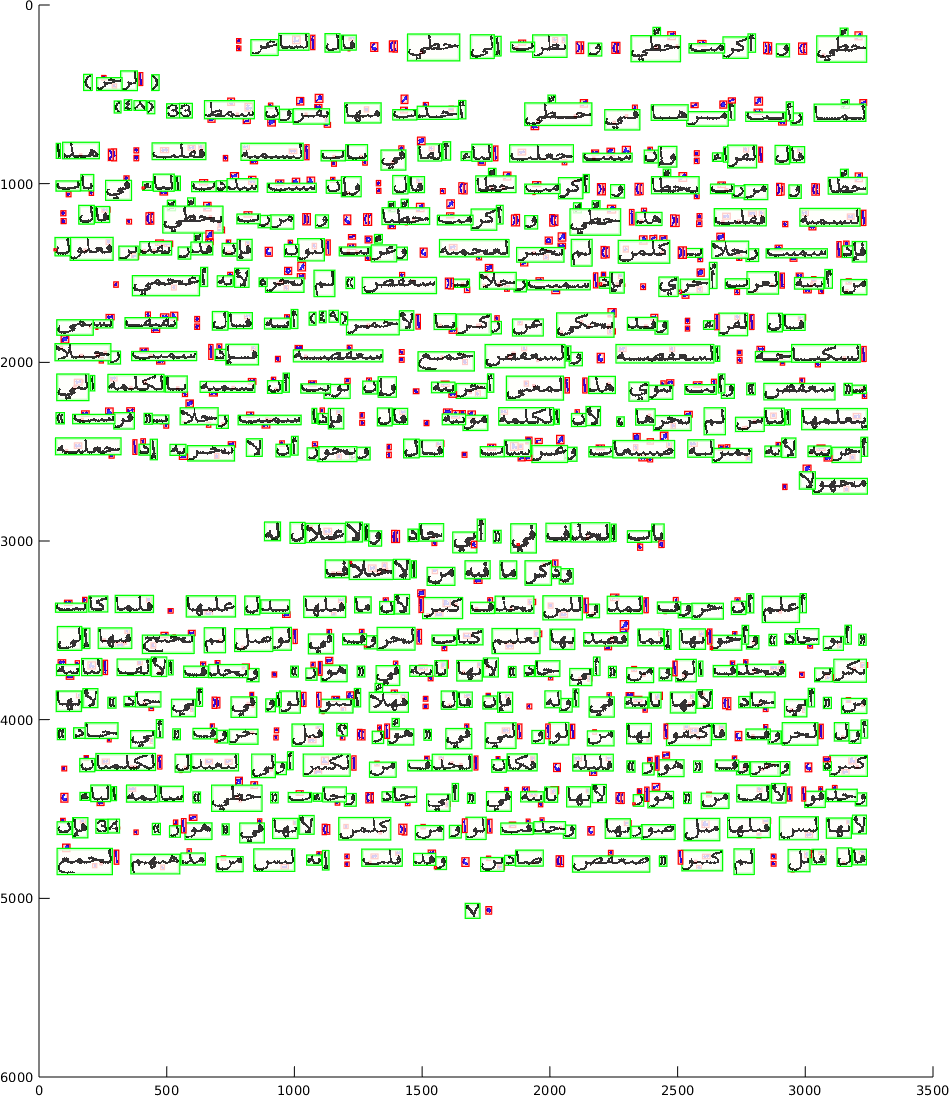}
    \caption{\label{fig:arabic-segmentation-output} The segmented page}
  \end{subfigure}
  \caption{\label{fig:arabic-segmentation}Arabic script
    segmentation. Large structures are enclosed in bounding
    boxes. Suspected punctuation and diacritics are detected.}
\end{figure*}

\begin{figure*}[htb]
  \begin{subfigure}[t]{0.5\textwidth}
    \includegraphics[width=\textwidth]{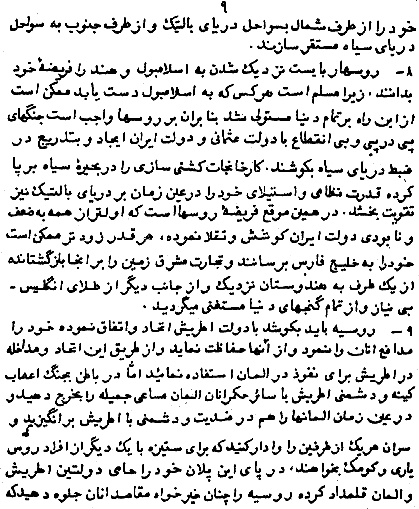}
    \caption{\label{fig:arabic-lines-of-text-input} A binarized sample page}
  \end{subfigure}
  \begin{subfigure}[t]{0.5\textwidth}
    \includegraphics[width=\textwidth]{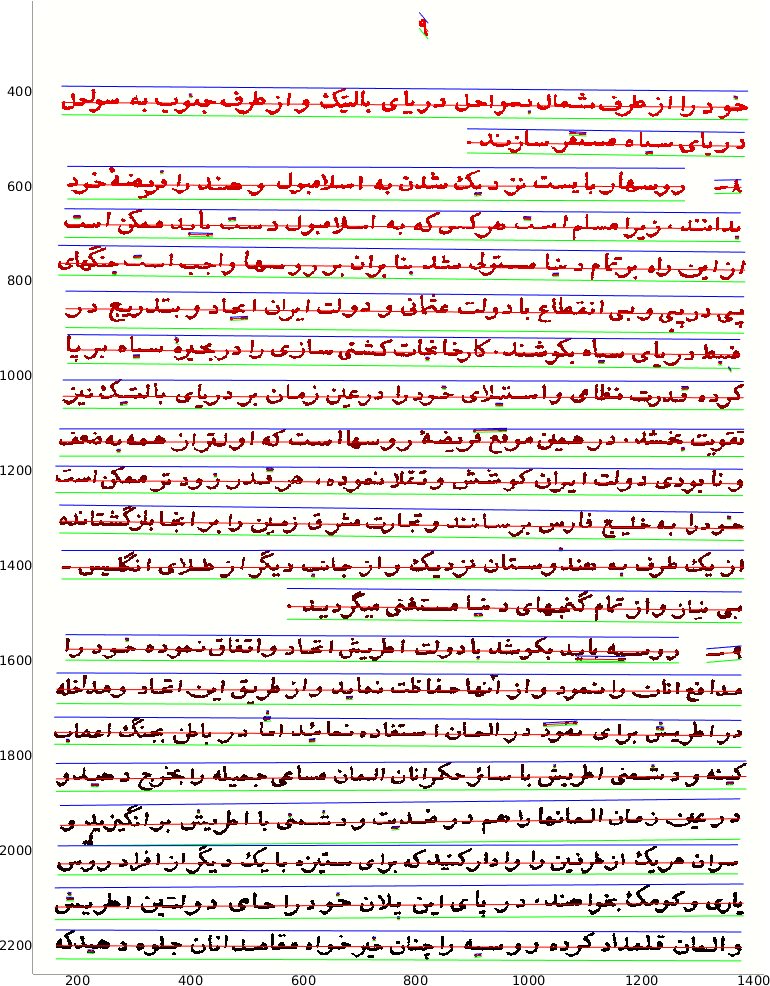}
    \caption{\label{fig:arabic-lines-of-text-output} Lines of text}
  \end{subfigure}
  \caption{\label{fig:arabic-lines-of-text}Determining lines of text
    and attachment of diacritics. Skewed and warped lines of text are
    detected by performing regression analysis on each line, computing
    3 auxillary lines (top --- blue, bottom --- green and middle --- red) which allow precisely
    determining the vertical position of characters in a line of text.}
\end{figure*}

In the current phase of the project our focus has been on
preprocessing of typical texts found in the ACKU dataset and similar
data. A major part of the preprocessing is page segmentation,
i.e., locating lines of text and separating large structures (ligatures
and words). Another aspect is correct attachment of diacritics, which
are a significant part of scripts based on the Arabic writing
systems. An example of segmentation performed by our software is found
in Figure~\ref{fig:arabic-segmentation}.  Another example of
segmentation is in Figure~\ref{fig:arabic-lines-of-text}, where lines
of text are identified very precisely, with diacritics and punctuation
properly identified.

\subsection{Traditional and Simplified Chinese}
\begin{comment}
  \sindy{"among others"? 2nd sentence below. I would delete}
  \marek{Went over the next paragraph, made numerous changes, issue addressed.}
\end{comment}
Chinese scripts are the second script class that we focus on. We
investigated several solutions to various subproblems. This includes
blob outline matching via correlations, dynamic time warping, and
broken character fixing. We have been investigating the use of
convolutional neural networks (CNN) in the recognition of isolated
characters. As a proof of concept, one of the datasets we are working
with is handwritten simplified Chinese characters. Our convolutional
neural networks investigate several input formats:
\begin{enumerate}
\item grayscale image;
\item binarized image;
\item transformed image data obtained by constructing \emph{blob
    outlines} (Figure~\ref{fig:chinese-page-outlines} and Figure~\ref{fig:broken-chinese-char}).
\item transformed image data obtained by \emph{skeletonization} (Figure~\ref{fig:skeletons});
\end{enumerate}
Generally, all methods tried work very well with Chinese
characters. We have not yet studied skeletonization extensively, so
this method will not be discussed extensively, but preliminary results and
literature search indicates that the method is useful and will be
studied further\cite{li_implementation_1993,khorsheed}.

Blob outlines are especially useful for large, historical Chinese
texts (e.g., Figure~\ref{Fig-5}), as they can be used for
\textbf{unsupervised learning}, i.e., grouping characters into
\textbf{clusters} (groups) of identical characters, only differing by
damage due to noise. It should be noted that for ancient documents the
fonts are unavailable; therefore clustering of characters can be used
to re-create the fonts for these characters. Outlines are somewhat
sensitive to character damage, and thus should be used in conjunction
with other techniques for superior accuracy.

We have positive (although considered preliminary) results with CNN as the neural
network when applied to grayscales and binarized images, showing
accuracies of up to 91\% with almost no preprocessing of the images,
when the class of characters is limited to 100.  We also conducted an
experiment of training a CNN on a set of 200 traditional characters
which were rendered in 56 fonts. When tested on 14 other fonts, a 97\%
recognition rate was found. The CNN-based method is one of the most
powerful methods known, and will be available in our software.

We are likely to use the \emph{committee approach} to Chinese, whereby
several methods are used to perform OCR, and the final decision will
be obtained by properly weighing the ``votes'' supplied by different
methods (committee members).  The committee approach is especially
useful for situations where the best method depends on the type of
document, and this is clearly the situation for the diverse body of
Chinese documents targeted by our project.

\subsection{Steps to developing OCR for the Bromello font}
\label{sec:steps-for-bromello}
This procedure has led to the development of an OCR system for a
specific Latin cursive script: Bromello \cite{bromello-font}. With small
changes, the procedure was applied to Farsi, and may be used as a template
to expanding the scope of our OCR system to new languages and scripts.

We started by downloading and installing on our Linux systems the
Bromello True Type Font (.ttf) file.  Then we generated a basic
training dataset consisting of the following sequences:

\begin{enumerate}
\item Unigrams, i.e., single letters, digits and punctuation
  characters.
\item Bigrams, i.e., all combinations of two characters. For
  Bromello, these sequences contain crucial information about how
  the characters connect.
\item Trigrams with a blank space in the middle. These sequences
  contain information about the inter-word space when words are
  close to each other.
\item Selected other trigrams. These were selected as a subset of
  those trigrams which were not properly decoded by the system
  trained on the above three types of training data.
\end{enumerate}

This resulted in approximately 12,000 short sequences. The results
included in this paper were produced by software trained for about
1 hour on a laptop computer.

Notably, we did not use dictionary words or any other linguistic
information (as many systems do).  Thus, the system is unaware of the concepts such as
word, sentence and grammar. Nevertheless, the system decodes the great
majority of English sentences with 100\% accuracy.

\subsection{Steps to developing OCR for Farsi and DejaVu Sans font}
By design, the procedure is very similar to the one in section~\ref{sec:steps-for-bromello}.
Thus, we will describe the differences. The alphabet consisted of all characters encountered in test data:
\begin{center}
  \arabicfontorimg{()*-.:[]،ءآأؤئابةتثجحخدذرزسشصضطظعغفقكلمنهوي٠١٢٣٤٥٦٧٨٩پچژکگی۰۱۲۳۵۶۷۸۹}{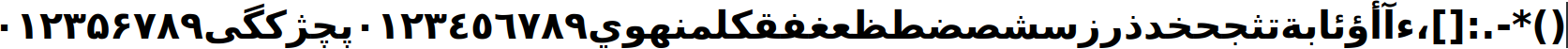}.
\end{center}

\begin{enumerate}
\item The Farsi alphabet was logically divided into 3 groups
  \begin{itemize}
  \item alphabetic (letters)
  \item numerals (digits); two sets of digits were identified in Unicode:
    \begin{itemize}
    \item Persian: \arabicfontorimg{۰۱۲۳۴۵۶۷۸۹}{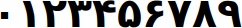}
      (Unicode range: 0x6F0--0x6F9);
    \item Eastern:
      \arabicfontorimg{٠١٢٣٤٥٦٧٨٩}{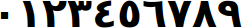} (Unicode range: 0x660--0x669)
    \end{itemize}
    as these are two different numeral systems encountered in test data;
  \item diacritics: \arabicfontorimg{ً َ ُ ِ ّ ْ ٔ}{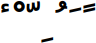}
  \end{itemize}
\item We included all unigrams in the training data, except for diacritics.
\item We included bigrams, excluding invalid ones (e.g., two diacritics).
\item We pre-trained the system with the training data above.
\item We included 1023 lines of Farsi from the OCR\_GS\_Data dataset.
\item We continued training on the entire dataset.
\end{enumerate}
We ran the training process for varying periods of time from 2 to 10
hours with various changes to the details. The 97\% accuracy on
training data was achieved with no more than 2 hours of training.
Pre-training on the basic unigram/bigram dataset significantly
shortened the training time.

The important factor in training on Farsi is the presence of samples
that exhibit all frequent medial forms, which requires placing
characters between two others, i.e. forming a trigram. As the number
of trigrams is prohibitively expensive (hundreds of thousands are
possible), the only medial forms that were seen by the system came
from the OCR\_GS\_Data dataset.

In automatic validation we encountered an interesting detail: false
negatives (errors that are not really errors), due to the fact that
\textbf{identical character glyphs are represented by more than one
  Unicode}. For instance, the two numeral systems (see above) overlap
in terms of the shape of the digits but two different Unicodes are
assigned to them.  It appears that even the human experts assigned the
one of the two possible Unicode codes without a clear reason.

%%% Local Variables:
%%% mode: latex
%%% TeX-master: "main.tex"
%%% TeX-engine: luatex
%%% End:

%% file: methods.tex
\section{Our methods and algorithmic basis}
\label{sec:methods}
As already mentioned, our approach is that of Machine Learning.  Both
for cursive scripts and for the logographic script of Traditional
Chinese we use an Artificial Neural Network (ANN) consisting of many
layers of artificial neurons (perceptrons). Generally, an ANN
consisting of more than 2 layers is considered a \emph{deep neural
  network} and the process of its training is known as Deep Learning
(DL). For cursive scripts we use Recurrent Neural Networks (RNN)
which look at the input as temporal data (time series). The timeline
is created by scanning lines of text left-to-right (Latin scripts)
or right-to-left (Arabic, Pashto and Farsi). In principle, the same
approach can be applied to Chinese, and our results indicate that
the accuracy is good. However, the main thrust of our effort
on Chinese was based on non-recurrent RNN.

\subsection{Cursive scripts and the video processing pipeline}
Cursive scripts are presented to machine learning software as
sequences of \emph{frames} which are several-pixel wide windows into
the data. Typically, this is an image representing a line of text,
obtained by preprocessing (cf. Figure~\ref{fig:arabic-lines-of-text}).
Thus, the sequence of frames is equivalent to a video sequence in
which we produce a panoramic view of a line of text to be converted
to Unicode.

The software needs to translate a long sequence of frames to a
relatively short sequence of characters. For example, in
Figure~\ref{fig:farsi-frame-1} the original image is 1416-by-93
pixels, the MATLAB-generated image is 750-by-53 pixels, and the Farsi
Unicode is 59 characters long. The mapping of 750 pixel columns to 59
characters is accomplished using a machine learning algorithm known as
Connectionist Temporal Classification (CTC) described by A.~Graves in
his dissertation \cite{graves_offline_2008} and has been subsequently
applied in several text-to-image conversion programs,
e.g., \emph{Kraken} and \emph{Tesseract} \cite{kraken,tesseract}. A
careful implementation of CTC in MATLAB has been important in our
approach as well. In section~\ref{sec:ctc} we provide further details
on CTC.

\begin{figure}
  \centering
  \includegraphics[width=0.5\textwidth]{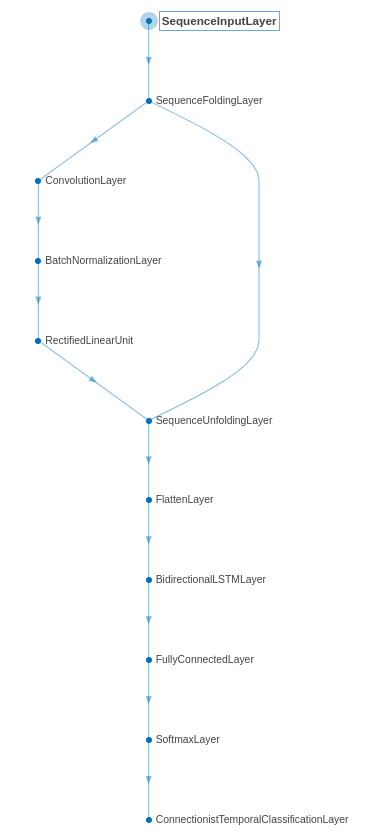}
  % TODO: Rename layers in the net so that this figure becomes more meaningful
  \caption{The layer graph of the DL network used in decoding cursive
    scripts.\label{fig:layers}}
\end{figure}

While the overall strategy of DL is dominant in OCR of cursive scripts,
the particular architectures differ in details. Figure~\ref{fig:layers}
displays our architecture, as depicted by a MATLAB tool. The reader
may note features typical of a video processing network. It is worth noting
that:
\begin{enumerate}
\item The workflow implemented by the network is that of
  sequence-to-sequence mapping (e.g., as opposed to image
  classification).
\item There is an convolutional layer portion of the network, enclosed
  by the sequence folding/unfolding layer.
\end{enumerate}
This architecture was chosen because it lends itself easily to
modification and experimentation, and is relatively easy to implement
in MATLAB. As OCR is performed on images of scanned pages, not on
video, there are two preprocessing elements which are applied before
submitting input to the DL network:
\begin{enumerate}
\item Images are segmented into lines of text. Several line-splitting
  algorithms have been produced by our project, and all of them have
  merits for specific types of content.
\item Lines of text are converted to video sequences using the sliding
  window approach, where a window of, say, 7 pixel columns is moved
  along the line of text, with a step of 1 pixel.
\end{enumerate}
Thus, the sample video for a 750-by-53 pixel input of
Figure~\ref{fig:farsi-frame-1} is converted to the video of 750 frames
of 53-by-7 video frames. All frames are binary images, as the original
image is binarized before processing, which is a customary step in
OCR.

\subsection{Blob outline analysis}
\label{sec:outlines}

In the current section we present a method we have developed for OCR on a
wide range of documents, with a substantial original research component.
Among others, the method solves the following problems:
\begin{enumerate}
\item Extremely accurate classification of characters; e.g.,
  near 100\% accuracy on Latin and Chinese characters of high and medium quality.
\item OCR on rotated and scaled text (cf. Figure~\ref{fig:rotated-text}).
\item Mapping one font to another, even if the fonts are quite distinct.
\item Accurate unsupervised classification of characters in an unknown font,
  e.g., grouping all 16,000 characters in an ancient Chinese book 
  utilizing fonts non-existent in modern typography.
\end{enumerate}

\begin{figure*}[htb]
  \begin{subfigure}[t]{0.5\textwidth}
    \includegraphics[width=\textwidth,height=2in]{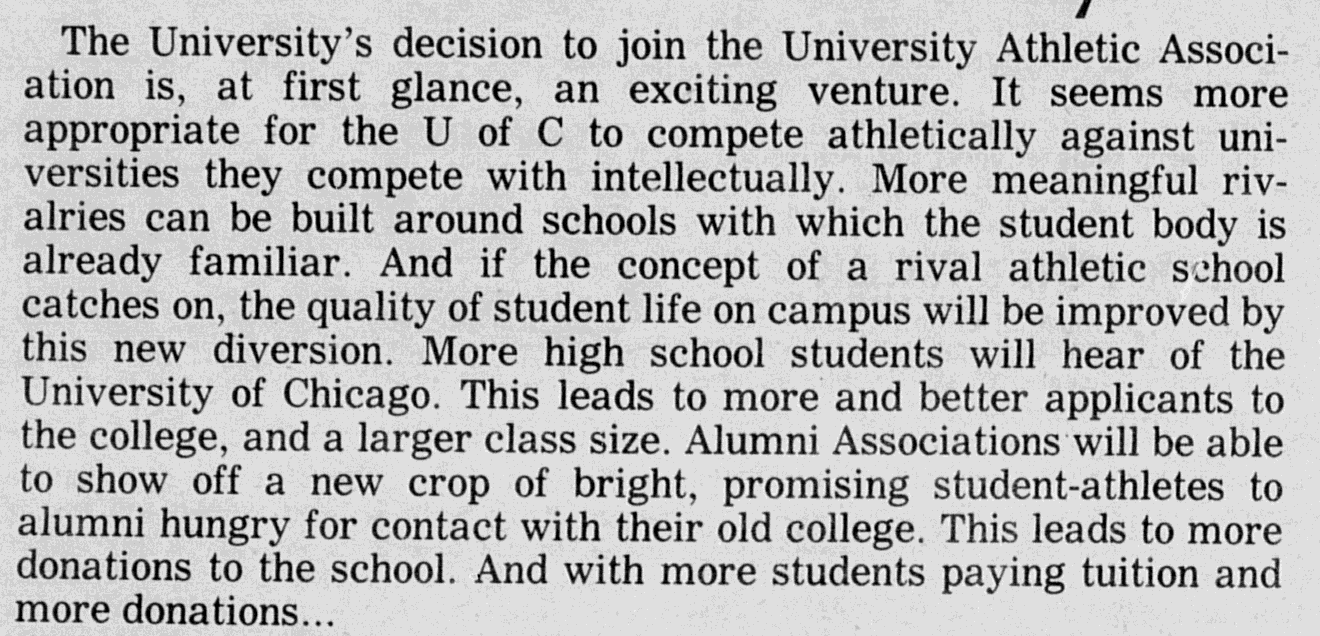}
    \caption{\label{fig:english-page-input} A campus newspaper paragraph}
  \end{subfigure}
  \begin{subfigure}[t]{0.5\textwidth}
    \includegraphics[width=\textwidth,height=2in]{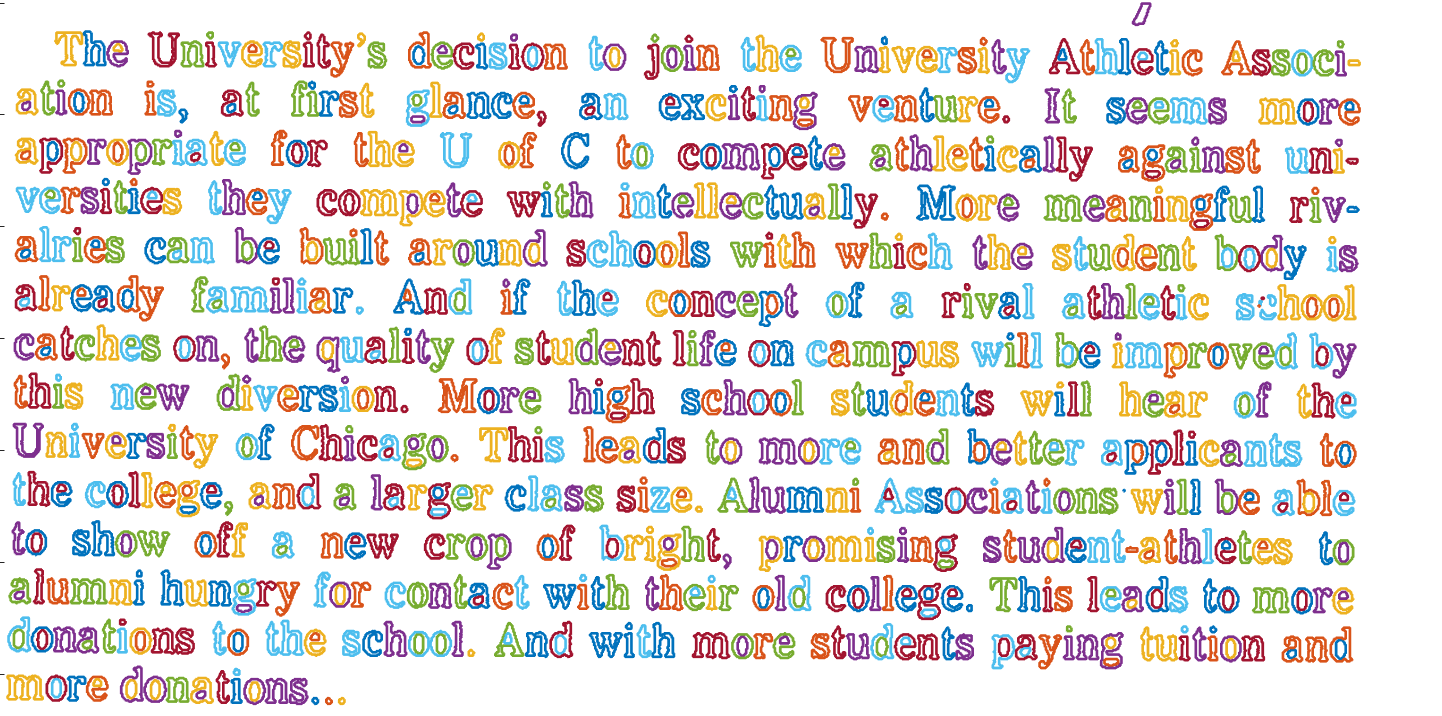}
    \caption{\label{fig:english-page-outlines-output}Outlines of all blobs}
  \end{subfigure}
  \caption{\label{fig:english-page-outlines} Blob outlines
    found for every character in an English newspaper paragraph.}
\end{figure*}

\begin{figure*}[htb]
  \begin{subfigure}[t]{0.5\textwidth}
    \includegraphics[width=\textwidth,height=4in]{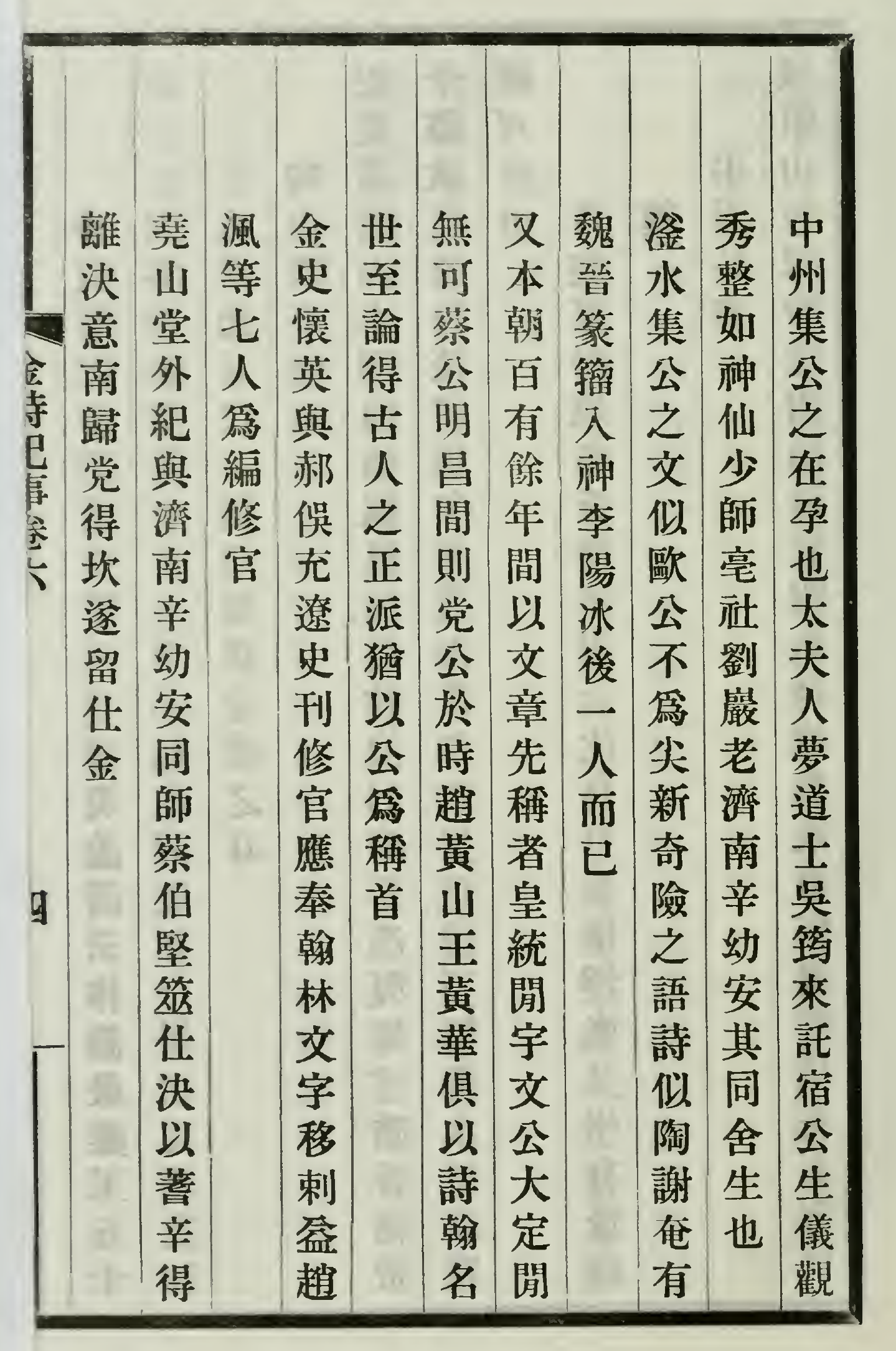}
    \caption{\label{fig:chinese-page-input} The original page}
  \end{subfigure}
  \begin{subfigure}[t]{0.5\textwidth}
    \includegraphics[width=\textwidth,height=4in]{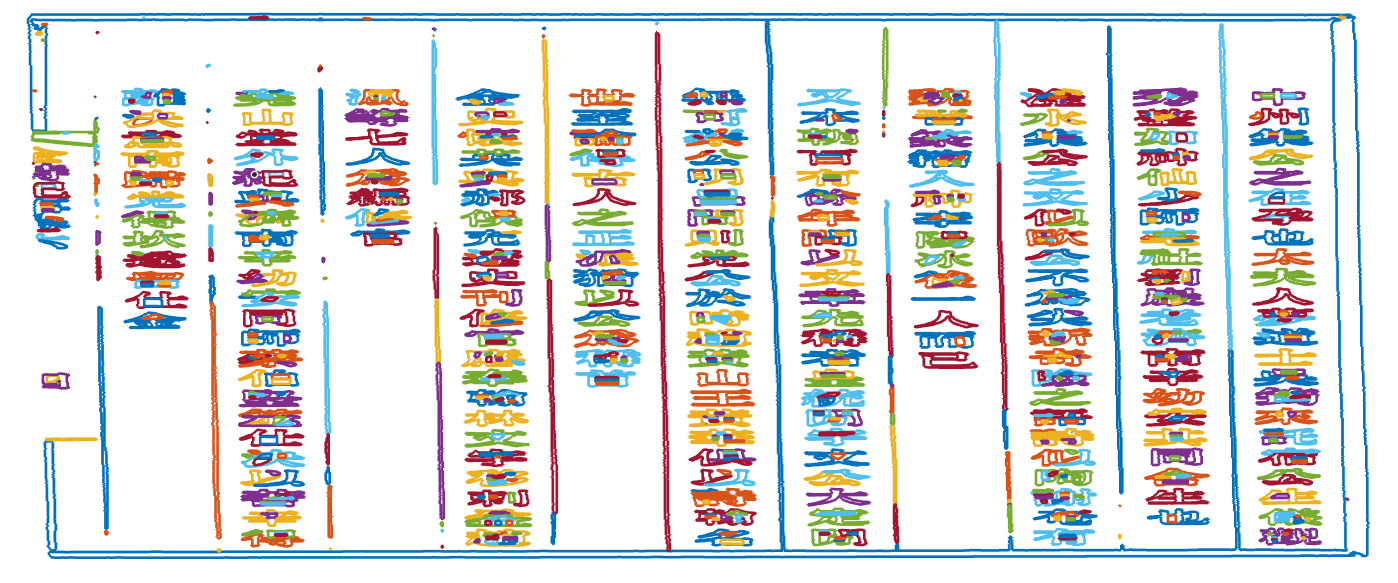}
    \caption{\label{fig:chinese-page-outlines-output}Outlines of all blobs}
  \end{subfigure}
  \caption{\label{fig:chinese-page-outlines}Character outlines
    found for every character in a Chinese book page.}
\end{figure*}

\subsection{The binary color model}
Scanned images for OCR are either grayscale or color. However,
conceptually print is black and white. Therefore, the color model of
image data for OCR uses only these two colors. Furthermore,
traditionally background is considered black (zero pixel) and writing
is considered white, or, more precisely saturated with maximum
brightness.  Digital scanners support this model, by lighting pages of
documents being scanned with extremely strong light, and therefore the
background is usually close to fully saturated, and characters are
close to black. In particular, the images produced by a digital
scanner are very different from images normally produced by a digital
camera in a cell phone, which is tasked with capturing all
intermediate brightness levels accurately. 

The conversion of color or grayscale input to binary color is called
binarization and is typically the first step in most OCR programs.
After binarization we normally work with the negative, i.e., we place
white characters on black background (cf. Figure~\ref{Fig-3}).
To avoid ambiguity the term \emph{foreground color} will always
refer to the color of the characters.

The software at this phase of processing has no idea of characters or
other semantic constructs associated with text. It only sees contiguous
areas of foreground referred to as ``blobs''. Most Latin characters 
consist of a single blob, with exceptions
such as `i', but Chinese characters
commonly consist of several blobs. For instance, in
Figure~\ref{Fig-3} we can see that the characters consist of (counting
from the left) 3, 2 and 2 blobs, respectively. It should be noted that
(semantically) the same character may be represented by a different
number of blobs, as a result of imperfections during the printing
process. In Figure~\ref{fig:broken-chinese-char} we have
an example from a Chinese book in which a character is either ``whole''
(consists of a single blob) or ``broken'' (consists of two blobs).

\subsection{Blobs and their outlines}
After conversion to black-and-white, a representation of a character
image may be extracted in the form of the outlines of the foreground
regions (the blobs).  These outlines consist of one or more
disconnected cyclical paths; for example, the Roman letter `f' has a
single outline cycle, whereas `e' has two (one inside the other).  A
larger sample of Latin character outlines is seen in
Figure~\ref{fig:english-page-outlines}. The approach also works well
on Chinese characters, as seen in
Figure~\ref{fig:chinese-page-outlines}.

We extract these outlines as oriented paths, so that the outside
boundary cycle can be distinguished from internal boundary cycles. We
developed two fast methods of extracting those outlines:
\begin{enumerate}
\item a graph-based approach, which identifies boundary points and
  then searches for loops that can be traversed;
\item a potentially faster line-sweep algorithm that scans across
  the image and simultaneously detects boundary points and joins them
  into loops.
\end{enumerate}
No information is lost in converting images to outlines--the image could
be reconstructed from the outlines. Nonetheless, the volume of data
is substantially reduced. Representing a character as a set of outlines
is a form of \emph{dimension reduction}, replacing the two-dimensional
image by a set of one-dimensional paths. For typical pages of Latin or 
Chinese text, the volume of data is reduced by a factor of 50.

After standardizing for position and character size, the geometric
distance between corresponding points on a pair of cycles serves as a
useful metric for comparing the cycles extracted from separate
character images, allowing character matching and grouping.

\marek{DTW was not used in the figure, but the projection method, in which
  the vertices of one outline get projected onto the vertices of one or many outlines.}

\begin{figure}
  \centering
  \includegraphics[width=.9\textwidth]{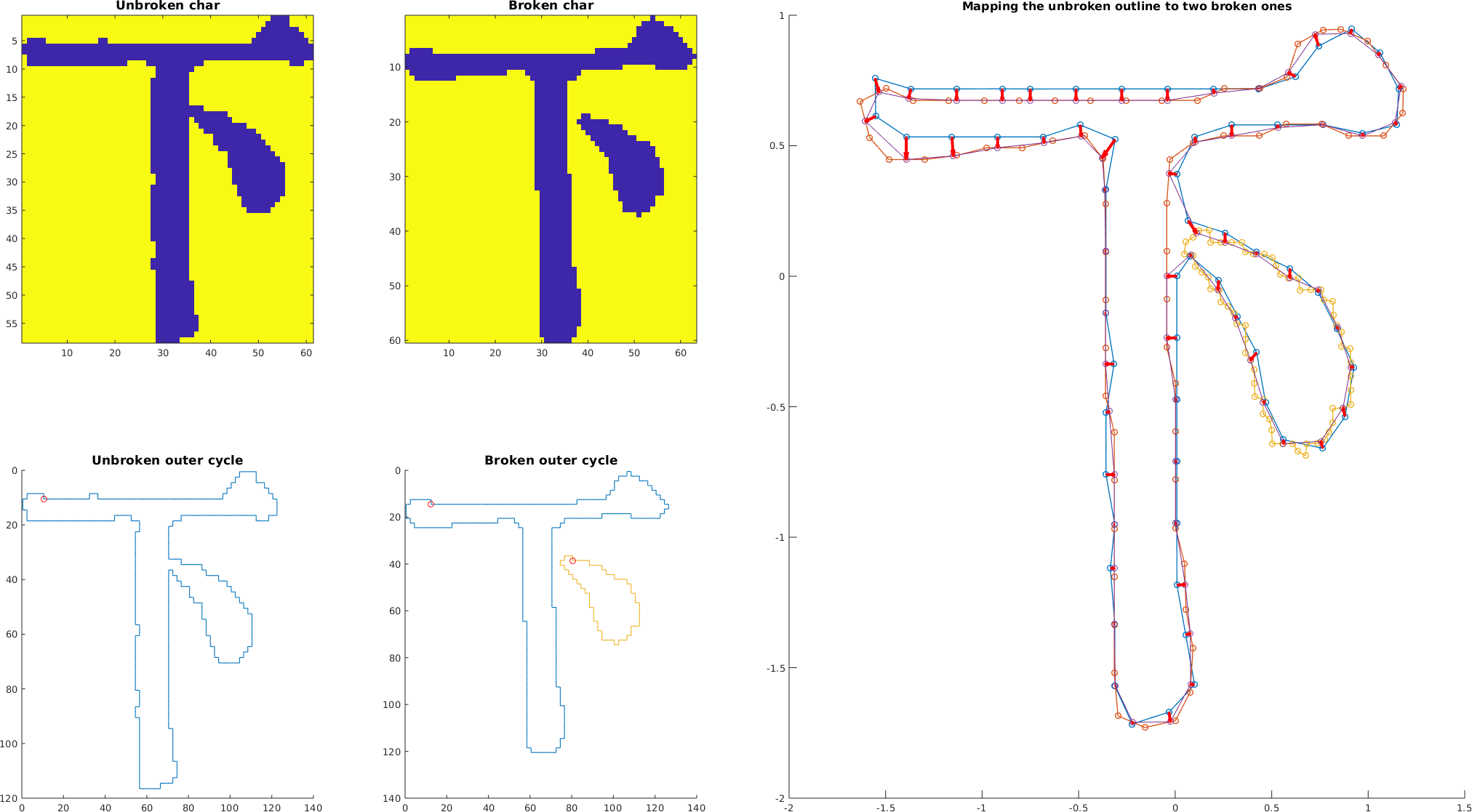}
  \caption{Matching outlines detects similarity between unbroken and
    broken versions of a Chinese
    character. \label{fig:broken-chinese-char}}
\end{figure}

A technique known as dynamic time warping (DTW) allows for matching
these ``corresponding points.''  In DTW, we consider each cycle as a
path traversed over time, and allow the traversing point on one cycle
to ``wait'' until the traversing point on the other cycle is as near
as possible before continuing to traverse.  This reduces errors due to
variation in starting position and minor noise or distortion of edges.
Furthermore, outline analysis may be used to match points on a pair of
cycles to points on a single cycle, allowing recognition of similarity
between two characters when one has been damaged (either by breaking
one boundary cycle into two, or merging two into one).
Figure~\ref{fig:broken-chinese-char} shows an example of this process,
where a character appears with a stroke either connected to the rest
of the character or disconnected.  The similarity of the outlines can
be detected despite the different connectivity.

In addition to distance-based matching and clustering, the extracted
outlines can be used as features for recognition by deep learning
algorithms such as the convolutional neural networks referred to in
the previous section.

\subsection{Skeletonization}
\begin{figure*}[htb]
  \begin{subfigure}[t]{0.5\textwidth}
    \includegraphics[width=\textwidth]{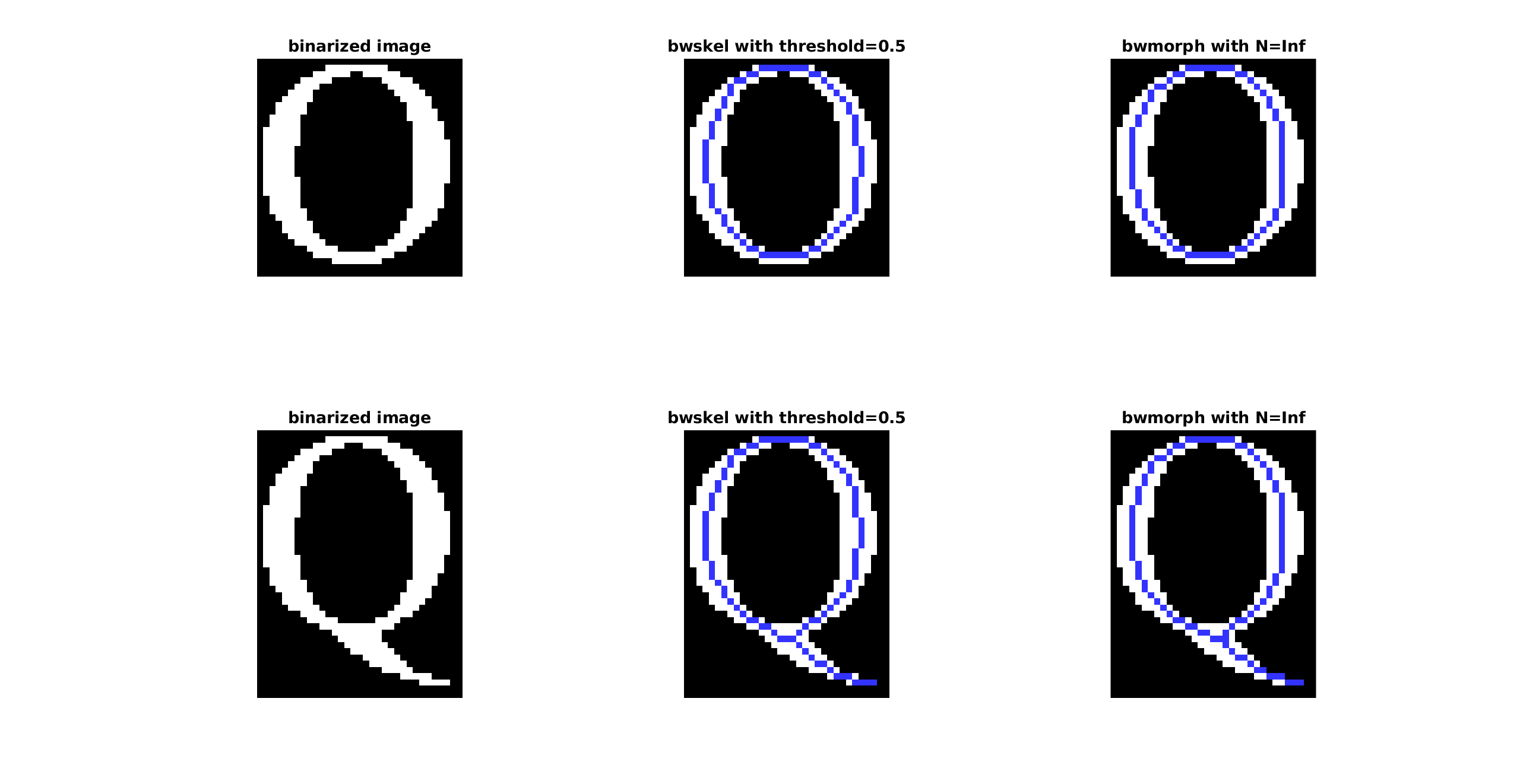}
    \caption{\label{fig:latin-skeletons}Skeletons of ``O'' and ``Q'' (in blue) }
  \end{subfigure}
  \begin{subfigure}[t]{0.5\textwidth}
    \includegraphics[width=\textwidth]{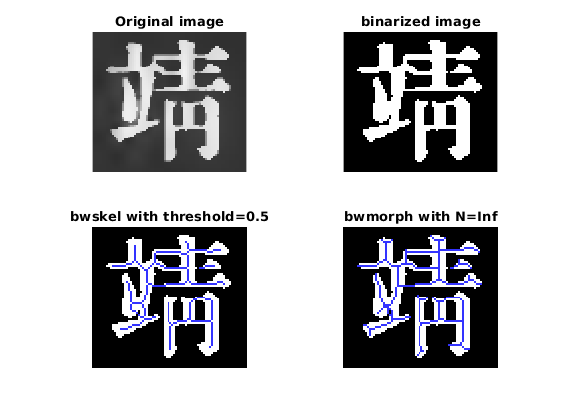}
    \caption{\label{fig:chinese-skeletons}Chinese character skeletons (in blue) }
  \end{subfigure}
  
  \centering
  \begin{subfigure}[t]{0.8\textwidth}
    \includegraphics[width=\textwidth]{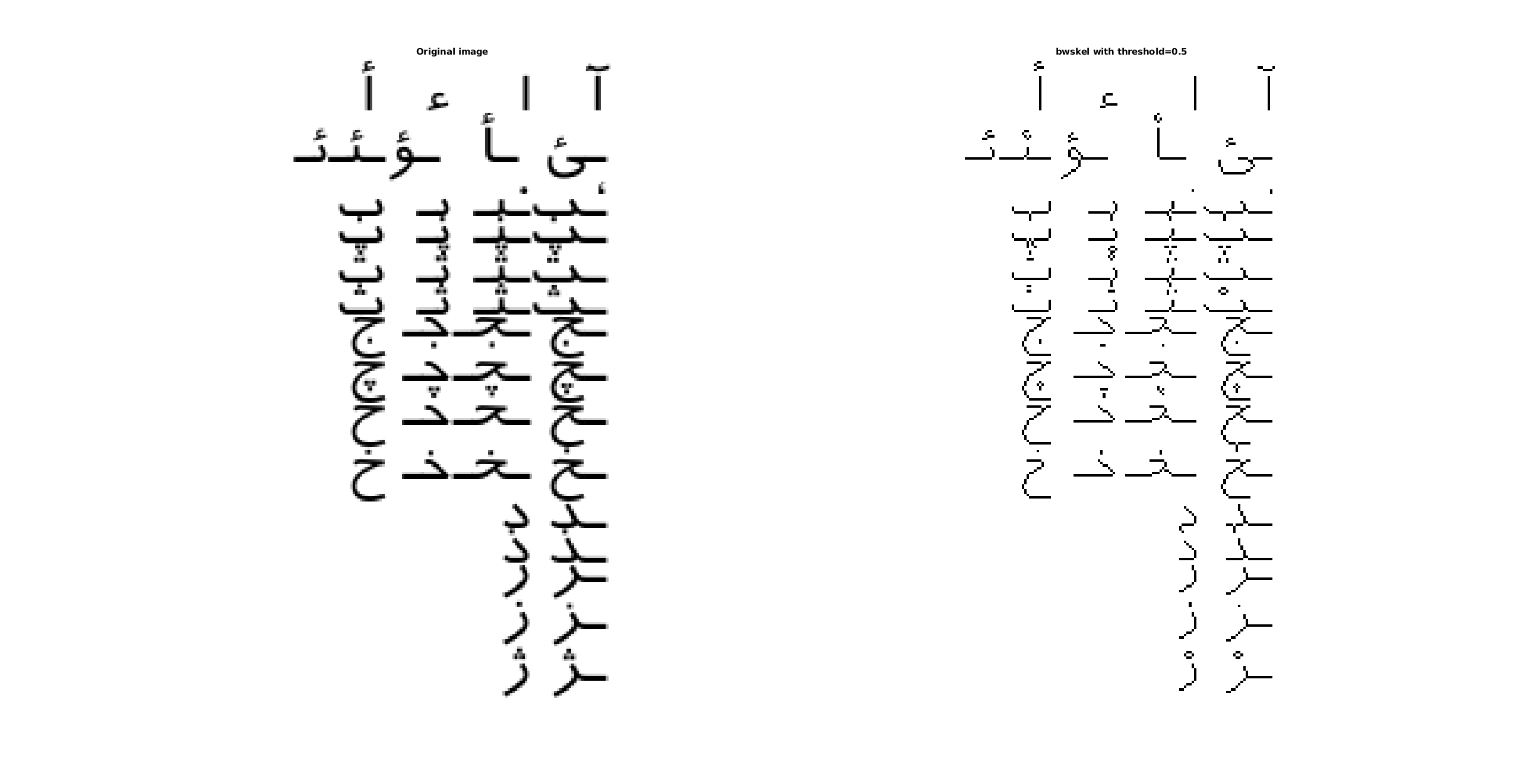}
    \caption{\label{fig:arabic-skeletons}Arabic character skeletons (right) }
  \end{subfigure}
  \caption{\label{fig:skeletons}Character skeletons obtained with two distinct
    MATLAB algorithms: 'bwskel' and 'bwmorph'.}
\end{figure*}

In the course of our research we researched skeletonization.
This technique converts characters to one-dimensional objects.
Like blob outlines (cf. section~\ref{sec:outlines}), it achieves
significant data reduction. Examples for Latin, Chinese and
Arabic scripts are found in Figure~\ref{fig:skeletons}. The essence of
the method can be described easily: characters occupying an area
of an image are thinned out to be one pixel wide, yet preserving
the shape of the original character.
Our preliminary research of this method indicates that replacing
characters with their skeletons does not lose the essential
information needed to convert them to Unicode. The advantage of this
method is that it opens new OCR approaches by bringing in a rich
supply of methods from graph theory (a branch of
mathematics). Skeletons are known to be useful in OCR. For example,
Ning Lin \cite{li_implementation_1993} reported an implementation of a
skeleton-based OCR system for Latin characters in 1991. Khorsheed
\cite{khorsheed} used skeletonization on his work on Arabic.

\subsection{Connectionist Temporal Classification (CTC)}
\label{sec:ctc}
\begin{comment}
\sindy{In following fix what's between the parentheses}
\marek{Rewritten extensively, so parentheses went away...}
\end{comment}
CTC applied along with RNN is considered the main advance in OCR of
cursive scripts, including Arabic and Farsi.  CTC has its origins in
\emph{speech recognition}. The original problem that it solves is the
fact that speakers time speech differently. Even the same speaker
sometimes speaks fast or slowly, or slurs certain vowels,
etc. Therefore mapping spoken language to written text involves
deciding the duration of various characters (or, more precisely,
\emph{phonemes}).  Therefore, the software needs to have a component
tantamount to a \emph{variable speed clock} which will match the speed
of the speaker with real time.  CTC provides such a clock, thus
automatically segmenting speech into its constituent elements.

It turns out that CTC can be applied to cursive scripts, to divide a
continuous stream of characters, connected according to complex rules
(or event without any discernible rules, as in handwritten text), into
constituent characters.  Hence, CTC solves one of the principal
difficulties in using a neural network (especially, a recurrent neural
network, or RNN) in OCR: the problem of \textbf{segmentation}.

The details of operation of CTC are beyond the scope of this paper,
but excellent descriptions can be found in blogs, and in Alex Graves's
Ph.D. dissertation \cite{graves_offline_2008}. Full understanding
involves knowledge of mathematics, and especially probability theory,
at the graduate level. However, to convey some more detailed understanding
of CTC we offer the following explanation: let us imagine reading a
line of text, viewing it through a narrow, vertical slit in a piece of
paper. We slide the paper from left to right with uniform speed (or
from right to left for Arabic) and try to read and understand the text
using this information. Our brain plays the role of RNN in this
case. Our memory allows us to remember a little bit of past
information (accounting for the word ``recurrent'' in RNN).  Our
neural network is trained to assign a probability of seeing a specific
character (or phoneme for speech) at any particular moment in
time. CTC works in tandem with the neural network by computing
\emph{the most probable timeline} for seeing various characters, thus
providing said variable speed clock.

It should be mentioned that in recent years the preferred neural
network to work with CTC is an RNN but in the past a simpler model
called a Hidden Markov Model (HMM) was used. RNN was demonstrated to
be superior.

\begin{comment}
  \marek{Not accurate, and complex. Replacing with my own.}

  The CTC segmentation approach begins by dividing a line of text into
  individual characters before passing the sequence of character images
  to the neural network for recognition.  This is a relatively easy
  problem when the script of interest consists of separated characters
  (e.g., printed text in languages using a Roman alphabet), although it
  can be complicated by the presence of noise, dirt, and other
  imperfections in the scanned image.  It is much more difficult to
  segment a line of text when the characters are connected to one
  another.
\end{comment}

% One of the key ideas behind CTC is that the neural network learns when
% letters transition to other letters without being told this in the training
% dataset. This allows for the network to be given datasets which are not broken
% out by character (i.e. unsegmented).

CTC sidesteps the necessity fo \textbf{manual segmentation} which was
sometimes applied in the past with great effort and time expense.
The segmentation problem was split between the neural network and CTC:
\begin{enumerate}
\item the neural network is focused on estimating
  the probability of each character as a function of time;
\item CTC is focused on adjusting the clock speed to produce
  the most probable sequence of characters.
\end{enumerate}

\begin{comment}
  the neural network to produce output at every horizontal pixel in the
  image.  In this way, the network associates to each horizontal
  position in the image an estimate of the most likely character present
  at that position -- an output known as a \emph{path}.  This is aided
  by augmenting the alphabet with a new ``blank'' symbol, $-$, allowing
  the network to represent the absence of character output at that
  position.  The result is typically a path that is longer than the
  correct labeling, because of some repeated characters and extra blank
  symbols.  After the image is processed, the path may be ``collapsed''
  down to a true labeling by removing all repeated outputs, and then all
  blanks, thereby producing a labeling without requiring that the
  network have advance knowledge of the boundaries between characters,
  or being able to explicitly detect such boundaries.  Although this
  appears at first to increase the computational complexity of the
  network, because many different paths may correspond to the same OCR
  output, an algorithm similar to that used in hidden Markov models
  allows for an efficient application of the CTC idea.
\end{comment}

We implemented CTC in MATLAB. Interestingly, no serious attempt at
MATLAB implementation of CTC is known to us, but we have found several
preliminary attempts, not good enough to be used in our system.  One
of the problems CTC encounters is a version of what is known as the
\textbf{vanishing gradient problem} in Recurrent Neural Network Theory
and first identified by Hochreiter \cite{hochreiter1998vanishing}.
The CTC version of this problem has to do with the necessity
to \textbf{multiply thousands of extremely small probabilities}
by each other. The problem is that we quickly
run out of digits of precision, when doing so, which
results in \emph{underflow}, i.e., the computer rounds
very small numbers to $0$, although they are not $0$.
Unfortunately, we need to calculate with such small numbers.

We believe that our MATLAB system addresses the problem of vanishing
gradients/underflow best of all CTC implementations known to us. It
should be noted that making this claim was not an objective, but it
was dictated by the demands of OCR on lines of Farsi. For example, the
bottom image in Figure~\ref{fig:farsi-frame-1} has dimensions
750-by-53, which results in $750\times53=43,500$ pixels, each pixel
either black or white.  
Thus, this particular image has probability of
occurring at random equivalent to obtaining \textbf{all heads} in
43,500 coin tosses! The probability of this is so tiny that no
computer can natively store it with any precision (the reader is
encouraged to calculate $0.5^{43,500}$ with her favorite
calculator). In principle, the solution to the problem is simple:
substitute probabilities with their logarithms.  The solution is
hinted at by many authors, including Graves, but our solution differs
in significant details. It is \textbf{guaranteed} not to produce
underflow for even the longest lines of text, in practice.

%%% Local Variables:
%%% mode: latex
%%% TeX-master: "main.tex"
%%% TeX-engine: luatex
%%% End:

%% file: software-overview.tex
\section{A brief overview of existing software and its limitations}
There are several software systems, both open source and commercial,
which potentially can perform text-to-image conversion on Farsi and
Pashto \cite{kraken,tesseract,abbyy_finereader_2019,sakhr}. However, we have
found that they are inadequate.  Mainly, the accuracy of these systems
is not adequate. As a rule of thumb, at least 90\% of the characters
of a document need to be recognized correctly in order for the OCR
result to be useful, e.g., to be incorporated in high-quality library
collections. While it has been reported that on some documents this
target number is met, the overall fraction of the ACKU dataset which
would meet the 90\% rule would be minuscule. It is known that the
success of OCR software on a particular document depends greatly on the
quality of the document and other content characteristics:
\begin{enumerate}
\item The image resolution.
\item The quality of the underlying media, e.g., paper warping, yellowing,
  chipping of paint, etc., generally called ``noise''.
\item Layout of the document. E.g., newsprint presents extra difficulty,
  due to its tabular nature, varying fonts and their sizes, and mixing
  of text and images.
\end{enumerate}
Therefore, software author claims of, say, 99\% rate of character
recognition are highly misleading without specifying exactly the
content on which the assessment was made. Despite the claim, a high
quality text may produce accuracy of 85\%.  On the other hand, we will
demonstrate that our software achieves 99\% or 100\% character
recognition rate on selected texts.

We would like to mention the project Open Islamicate Texts Initiative
(OpenITI) utilizing the program \emph{Kraken} \cite{kraken}, which is
similar in scope to our project. The project produced high quality
reference data \cite{ocr-gs-data}, and a white paper accurately
summarizing the software performance on carefully curated test data
\cite{romanov2017important}. In our current project we utilized the
OpenITI dataset to test our own software. What is different about our
project is the emphasis on research, including development of new
algorithms and workflows, which will substantially expand the class of
documents for which the 90\% accuracy threshold is met.

\begin{figure}
  \centering
  \includegraphics[width=0.8\textwidth]{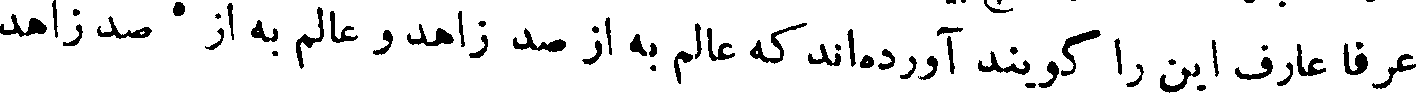}
  \includegraphics[width=0.8\textwidth]{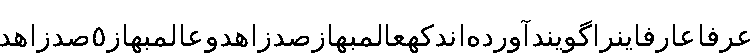}
  
  \caption{\label{fig:farsi-frame-1} A line of Farsi from the
    OCR\_GS\_Data dataset. The top line is the original image. The
    bottom is an image rendered from the Unicode label of the image
    with MATLAB.}
\end{figure}

%%% Local Variables:
%%% mode: latex
%%% TeX-master: "main.tex"
%%% TeX-engine: luatex
%%% End:

%% file: yan.tex
\section{Evaluation of other OCR software}
\begin{comment}
  \marek{Someone fix Figure Numbers to be consistent with Yan docx}
  \dwight{Apparently someone fixed it since it seems fine to me}
  \yan {I fixed the figures}
\end{comment}

We start with a disclaimer: a fair evaluation of complex software
systems that OCR systems are is a nearly impossible task. We evaluated
open source and commercial systems. Some systems were not available to
us due to high cost.  Open source systems are too complex to be
completely understood. Moreover, software this complex is controlled
by a myriad of parameters which affect its operation in often dramatic
ways. Thus, \textbf{the accuracy can be greatly affected by the way we
  prepare data}. It would appear that a fair test would be to present
the same document to two software systems, and see which produces more
accurate results.  However, this offers an unfair advantage to commercial
systems which do not necessarily have better OCR algorithms, but
devoted much more effort to coverage of different types of content.

\marek{I must write up my Tesseract experiment on Chinese newspaper,
  which increases accuracy to above 90\% by presenting the program
  with a region of a page, rather than the entire page}

Thus, our results are typical of what one would obtain from the
respective software system with minimum intervention and tweaking of the
parameters to produce better results.

In our studies, multiple commercial and open-source OCR systems and/or
online services have been evaluated for their accuracy rates on three
languages with four writing scripts (Simplified Chinese, Traditional
Chinese, Persian, and Pashto) from 2017 to 2020 for several
purposes. These purposes include:
\begin{enumerate}
\item to measure the accuracy of these systems, specifically
  for these languages;
\item to gain an understanding of their architecture and their underlying models;
\item to benchmark their performance;
\item to assess their trainability, using random testing data.
\end{enumerate}
The primary focus of these tests is on character and word accuracy
rates. Other factors such as pricing and running time are not
important for the purpose of our research at this stage.

\subsection{Issues with reported OCR accuracy}
A recent evaluation by Tafti et.al. of multiple OCR systems
concluded that Google Docs and ABBYY performed better than others \cite{tafti_ocr_2016}.

Readers should be aware that there are underlying issues when
evaluating these systems for their accuracy rate. Besides common factors
such as quality of document images and background noise, there are
other factors that affect the OCR quality greatly. Often we see
reports showing very high accuracy rates for certain documents. We
have observed another set of documents come with inconsistent OCR
quality, given the difference in terms of retrospective publication
dates and complex layout. As a result, the accuracy report might be
subjective to a specific set of documents and languages.  Our
experience shows that there are several critical factors when
evaluating accuracy rates: 

\begin{trivlist}
\item[ \textbf{Language:}] Typical OCR systems provide over 100+
  languages support. However, one must understand that languages are
  complex subjects and there exists some fundamental differences in
  languages, from structure to writing order. 
  For example,
  \begin{itemize}
  \item Retrospective Traditional Chinese is written vertically, in
    columns, from top-to-bottom, columns ordered right-to-left;
  \item Current Traditional Chinese is predominantly 
    written horizontally, in rows, characters in each row ordered from left-to-right,
    and rows ordered from top-to-bottom.
  \item Sometimes vertical Traditional Chinese writing
    is still used in modern books.
  \item Arabic, Persian/Farsi, Pashto, and Urdu are written
    horizontally, in rows, with characters in each row order from
    right-to-left and rows ordered from top-to-bottom.
  \end{itemize}
  
\item[ \textbf{Cursive and non-cursive:}] Arabic alphabet and
  adaptations of the Arabic alphabet such as Persian and Pashto are
  written from right-to-left in a cursive style, and letters vary in
  shape depending on their position within a word. Letters can have
  up to 4 distinct forms corresponding to an initial, medial, final,
  or isolated position.
\item[ \textbf{Layout:}] Documents with simple and straightforward
  layout tend to allow OCR systems to produce very high accuracy,
  while newspapers with complicated layout mixing figures in
  multiple irregular columns typically bring low-quality (and sometimes
  unusable) results.
\item[ \textbf{Fonts:}] Documents with unusual fonts typically will
  receive bad OCR results. It is hopeful that a new generation of OCR
  with CNN or LSTM will produce better OCR results.  Our effort is to
  evaluate OCR systems specific to languages. We have found that
  certain OCR systems work better for certain languages and certain
  OCR systems work better related to layout.
\end{trivlist}
  
We favor open source systems over commercial systems, because these
systems can be ported to multiple devices and can be trained for
various applications. As an example, we have successfully conducted
training of \emph{Tesseract}, and we build simple applications based
on the \emph{Tesseract} API (Application Programmer's Interface).
Training \emph{Kraken} is also straightforward, and the source code
base is small and well-written, in Python. It is easy to understand
and modify. The successes of open source systems are readily tracked by
the community contributions to these systems by non-developers, in the
form of \textbf{trained models} and \textbf{training data}, which
extend the coverage of various languages.  In addition, these systems
come with pre-trained models for many languages and are shipped with
training data.

\marek{I could not figure out why this was mentioned:
The (e.g. Raspberry,
mobile, cloud services).
}

\marek{I don' like the next sentence. We already have the testing
  data in the repository; we should be able to say that
  the data are in the GitHub public repo.}
\yan{I agree to say that the testing datasets are available in the github. }
Our testing datasets are available in the Github public repository. 

\subsection{A list of tested OCR systems}
\subsubsection{Tesseract}
\begin{comment}
\sindy{I would cut out some of this about Smith stuff. Redundant and not relevant.}
\marek{I shortened Ray Smith bio.}
\end{comment}
Tesseract is probably the most well known open-source OCR software. It
was originally developed at HP Laboratories Bristol and HP at Greeley
Colorado from 1985 to 1994. In 2005, Tesseract was open sourced by HP,
and it has been sponsored by Google since 2006.  The leading developer
has been Ray Smith, starting from the inception at HP Labs
in 1987, until today.

\begin{comment}
, who received his Ph.D. in Computer Science,
focusing on OCR. He was developing the Tesseract OCR engine at HP Labs
Bristol for 10 years starting in 1987, then 3 years related work at HP
Greeley Colorado for HP's PrecisionScan product, and 7 years
developing the Omnipage OCR from 1998-2005, and work at Google on
Tesseract again in 2005.  
\end{comment}

One of the most highly cited papers related to Tesseract was written
by Ray Smith \cite{smith_overview_2007}.  However, readers should be
aware this paper was published in 2007, before RNNs and specifically
LSTM, gained entrance into the OCR arena. Therefore, the paper does
not cover the LSTM implementation in version 4.0. We have tested
multiple versions of this system, including:
\begin{trivlist}
\item[\textbf{Version 3.0:}] First version 3.0 released in July 2015. The last
  patch release was 3.05.02, and was released on June 19, 2018. It
  added more language support than the previous version, including
  language models for traditional Chinese (chi\_tra and chi\_tra\_vert),
  Persian and Farsi (fas) and Pashto (pus).
\item[\textbf{Version 4.00.00 alpha:}] Released in Oct 2018. This was the first
  version implementing LSTM, and it is the the same 
  RNN as is used in \emph{Kraken}.

%\sindy{"still" not satifactory? versus what?}

\item[\textbf{Version 4.1.1:}] This version was released in Dec 2019. Version~4
  adds an LSTM machine learning OCR engine. We have tested this in the
  official trained data version and the version with our own trained
  data. However, the result was still not satisfactory as compared
  to the leading systems.
\end{trivlist}

\subsubsection{Google Cloud and Google Drive}
Google Cloud and Google Drive platforms offer OCR services. The
documentation is available at
\url{https://cloud.google.com/functions/docs/tutorials/ocr}. Given that Google
has been supporting Tesseract since 2005, and Tesseract lead developer
Ray Smith has been working at Google and Google Deepmind since
2005, it is reasonable to believe that Google uses Tesseract, but we
cannot verify this. Walker compared Google Cloud and Tesseract in 2008,
and showed that Google performed better than Tesseract in Arabic,
English, Hindi, Japanese and Russian \cite{walker_web-based_2018}.

\subsubsection{ABBYY}
ABBYY has been in the OCR business since 1989 specializing in OCR
technology. Its products have been licensed to many scanning and
document solutions companies such as Fujitsu, plustek and
Panasonic. Many universities have licensed its products.
\begin{itemize}
\item  Finereader 11 (windows) and 12 (Mac). 
\item  Abbyy.cloud: ABBYY's cloud service. It is straightforward to set up. 
\end{itemize}
  
\subsubsection{Convertio}
An online service is provided by Softo Ltd. It has a 4.7/5 high
overall conversion quality rating.

\subsubsection{NovoVerus}
The software provider is NovoDynamics; NovoVerus Professional is
\$6,500 USD plus an annual \$1,300 USD license.

\subsubsection{Sakhr}
On its website, Sakhr states that it is a pioneer and market leader in
Arabic language technology and solutions with 28+ years of research
and development. It claims that its OCR technology
(\url{http://www.sakhr.com/index.php/en/solutions/ocr}) was ranked \#1
by U.S. government evaluators, but we cannot find any evidence to back
this claim. With email exchanges with its sales representative, the
company claimed to have up to 99\% accuracy depending on the image
quality (300 DPI) for Persian/Farsi and Pashto if their fonts are
Arial or Simplified Arabic. Unfortunately, the company required us to
pay \$1300 upfront without giving us a chance to evaluate their
product.

\subsubsection{Kraken}
\emph{Kraken} \cite{kraken} is an OCR system written in Python.  It
features a small code base of approximately 8,000 lines of Python
code\footnote{An automated count of kraken, version 2.0.8, yields
  7,091 lines of Python code.}.  However, the small line count
may be misleading, as \emph{Kraken} relies upon a vast code base
available to a Python developer in a myriad of Python libraries.  Some
of the most critical lbraries include \emph{clstm} and \emph{pyrnn}
for RNNs, a massive ML library \emph{torch}, a scientific computing
and math library \emph{scipy} which is meant to provide functionality
similar to MATLAB. Kraken is currently focused on cursive scripts
(mainly Arabic) and does not provide a model for Chinese.

\subsection{Testing on Traditional and Simplified Chinese}
Simplified Chinese characters
\chinesefont{简体字}
are standardized Chinese characters for official use in mainland
China, Singapore and Malaysia. Professor Qian XuanTong, a linguist,
proposed to simplify Chinese characters in 1920. With growing
interests, the Republic of China Department of Education in August
1935 released the first table of simplified Chinese Characters
containing a total of 324 characters, which have been used in the
society for a long time. This act was retracted in February 1936 due
to strong opinions from some government officials. In 1952, the
government of People's Republic of China (mainland China) started to
implement Simplified Chinese characters, and by 1964 the Table of
General Standard Chinese Characters was released.

Traditional Chinese characters have over 3,000 years of history. They
are still officially used in Hong Kong, Macau, and the Republic of
China (Taiwan). As one of the major characters in the world, it has
influences on other character sets such as Korean Hanja and Japanese
Kanji. It is estimated that over 2,000 Traditional Chinese characters
are still used in Kanji. Traditional Chinese Characters and simplified
Chinese characters are one of the two standard character sets of the
current Chinese written language system.

In a broad sense, simplified Chinese characters are simplified
versions of traditional Chinese characters in terms of ``structure'',
some of which have existed for thousands of years along with their
regular (more complicated) forms. Simplified characters were created
by reducing the number of strokes.

\begin{figure*}[htb]
  \begin{subfigure}[t]{0.5\textwidth}
    \includegraphics[width=\textwidth]{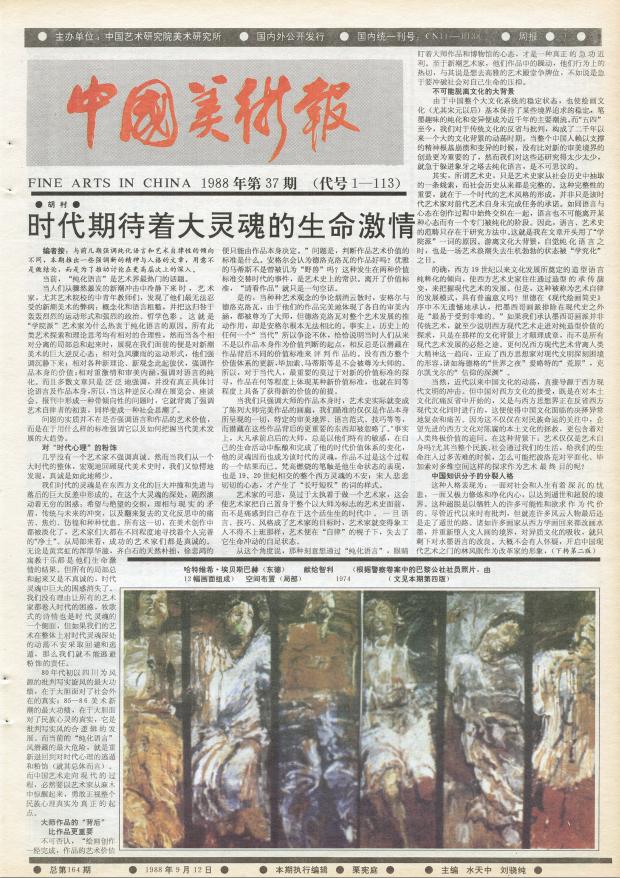}
    \caption{\label{Fig-1} Simplified Chinese Newspaper}
  \end{subfigure}
  \begin{subfigure}[t]{0.5\textwidth}
    \includegraphics[width=\textwidth]{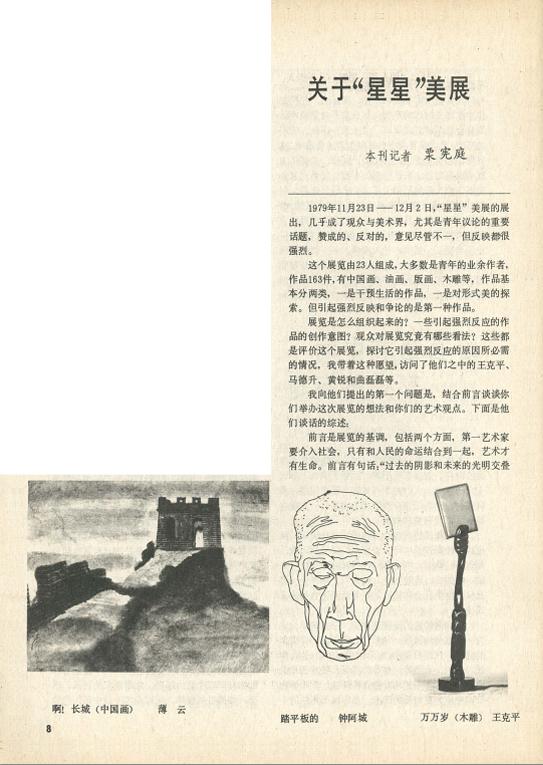}
    \caption{\label{Fig-2} Simplified Chinese Book}
  \end{subfigure}
  \caption{\label{Fig-1-and-2}Chinese Newspaper and Simplified Chinese Book}
\end{figure*}

\section{The Tests}
\subsection*{Simplified Chinese}
\subsubsection*{Tesseract 3.0}
\begin{itemize}
\item This version achieved 90\%+ accuracy rate if the paper was 
  simply a text layout without any images, columns or variation in background intensity.
\item Fig.~\ref{Fig-1} and Fig.~\ref{Fig-2}: we recorded 30\%-60\%
  accuracy rates depending on the degree of complexity of the layout or background (Tested
  January 2020).
\end{itemize}

\subsubsection*{ABBYY Fine reader 11 (Windows) and 12 (Mac)}
\begin{itemize}
\item Fig.~\ref{Fig-1} and Fig.~\ref{Fig-2}: ABBYY generated
  accurate page segmentation as well as a higher accuracy
  rate of 86\%+ (Tested May 2019).
\end{itemize}

\subsubsection*{Adobe Acrobat Pro DC}
\begin{itemize}
\item Fig.~\ref{Fig-1}:
  Page segmentation was not 100\% correct. Character accuracy rate was 83\% (Tested March 2020). 
\item Fig.~\ref{Fig-2}: Page segmentation was in order and the rate of character accuracy was 84\%. (Tested March 2020)
\end{itemize}

\begin{figure}
  \centering
  \includegraphics[width=0.99\textwidth]{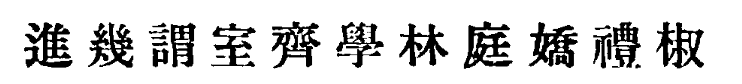}
  \caption{\label{Fig-3}: Eleven Traditional Chinese Samples}
\end{figure}

\begin{figure}
  \centering
  \includegraphics[width=0.5\textwidth]{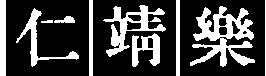}
  \caption{\label{Fig-4}: Three Traditional Chinese Samples}
\end{figure}

\begin{figure}
  \centering
  \includegraphics[width=0.9\textwidth]{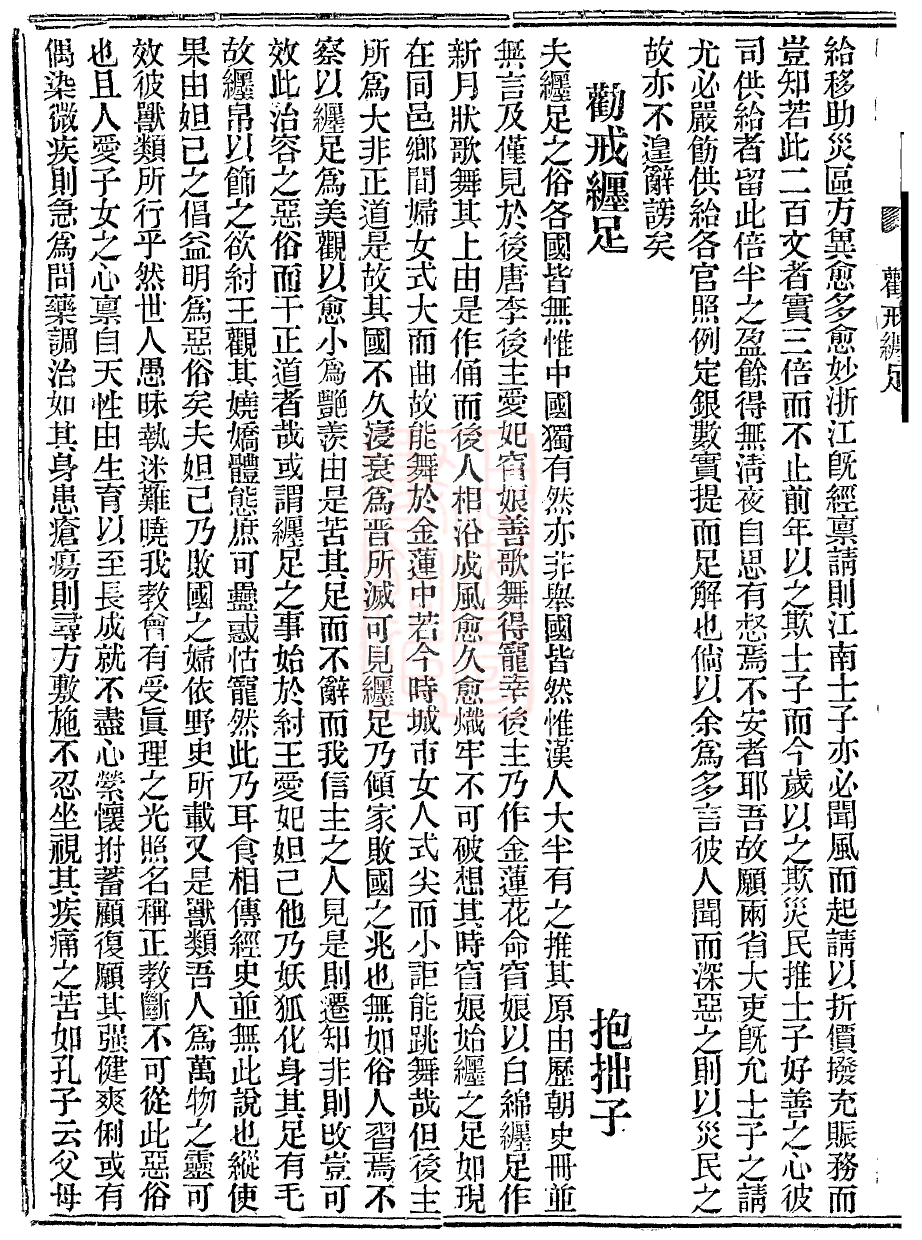}
  \caption{\label{Fig-5}: A Traditional Chinese Monograph}
\end{figure}

\begin{figure}
  \centering
  \includegraphics[width=0.99\textwidth]{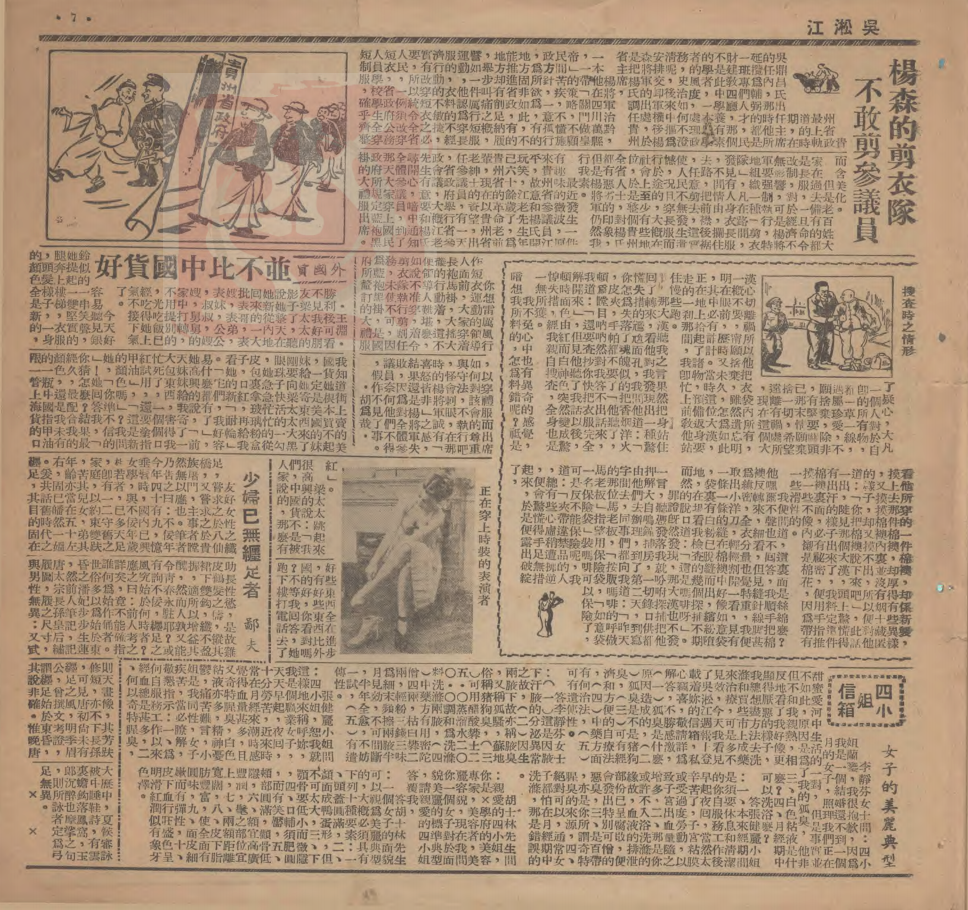}
  \caption{\label{Fig-6}: A Traditional Chinese Newspaper. This sample
    features a complex layout for which no software succeeded at producing good OCR results.}

\end{figure}

\subsection{Traditional Chinese}
\subsubsection{Tesseract}
\begin{itemize}
\item{Version 4.00.00alpha:}  Fig.~\ref{Fig-3}: We fed 11 randomly selected characters to the
  following OCR systems. Accuracy rates were very good. Tesseract and
  Convertio achieved 91\% accuracy rate, while O2CR achieved 82\% (
  Tesseract the 4th character, O2CR the 9th and 10th character,
  Convertio the 9th character) (Tested May 2019)

\item{Version 4.11}: Fig.~\ref{Fig-4}: We fed 100 random characters (selected from
  our GitHub repository, see Fig.~\ref{Fig-5} for samples) and
  recorded 60\% accuracy rate (version 4.11) vs. 36\% (version 3.0)
  (note: version 4.11 with pre-trained data).  After training using
  our 100 pic dataset, the result improved, but not as much as we
  wanted. (Tested Feb 2020)
  
  There might be two reasons for this result.
  \begin{enumerate}
  \item  A larger training dataset may be necessary.
  \item Characters might need better quality. We will run more tests for this version
  \end{enumerate}
\end{itemize}
  \sindy{Missing word about segmentation follows}
  \marek{Can't address this.}
\subsubsection{Google Drive (Convert PDF and photo files to text)}
\begin{itemize}
\item Fig.~\ref{Fig-5}: The result was good. It was very time consuming
  compared to a local system. We have to wait 30 minutes to get one page
  of PDF to be converted.
  \marek{Could not fix sentence about page segmentation.}
  %Page segmentation is .
  (Tested March 2020)

\item Fig.~\ref{Fig-6}: No result was returned after a four hour wait time 
  for the page to open in Google Doc. (Tested March 2020)
\end{itemize}

\subsubsection{ABBYY }
FineReader 11 (Windows) and 12 (Mac):
\begin{itemize}
\item Fig~\ref{Fig-5}: The result was very disappointing. Page
  segmentation was wrong. (Tested May 2019)
\item Fig~\ref{Fig-6}: The result was disappointing.  (Test May 2019)
\end{itemize}
  
\subsubsection{Abbyy.cloud}
\begin{itemize}
\item  Fig 3: It produced a high accuracy rate of 91\%. (Tested March 2020)
\item Fig.~\ref{Fig-5}: High accuracy rate of 90\% (Tested April
  2020). It is slightly better than Adobe Acrobat Pro DC (see below),
  but produced different errors including punctuation. Please note
  that old Traditional Chinese materials typically does not have
  punctuation.

\item Fig.~\ref{Fig-6}: No output was produced and an incorrect error message was issued. (note: The
  website returned message ``was not processed: the recognized
  document contains errors''). (Tested March 2020)
\end{itemize}

\subsubsection{Adobe Acrobat Pro DC}
\begin{itemize}
\item Fig.~\ref{Fig-5}: We have the same PDF file digitized from the
  original monograph in color, and generated three file formats: A PNG
  file (exported from Acrobat Pro DC from page 1 of the PDF with 600
  DPI in color), A TIFF file (exported from Acrobat Pro DC from page 1
  of the PDF with 600 DPI in color) and a PDF file (screenshot of page
  1 of the original PDF with estimated 88 DPI from a 49 inch 4K
  Monitor). We received different results with the same software. The PDF
  generated a good accuracy rate of 83\% , while Acrobat Pro DC
  displayed ``Acrobat could not perform Text Recognition on this page
  because: Unknown error'' for both the PNG file and the TIFF file
  generated. We reduced both PNG and TIFF images to its 33\% and
  re-ran OCR. We received different OCR results. The lower resolution
  of screenshot had slightly better results (Tested March 2020) and
  both PNG and TIFF images had the same OCR outputs.
\item Fig.~\ref{Fig-6}: Output was not recognizable and page
  segmentation was faulty. (Tested March 2020)
\end{itemize}

\subsubsection{Convertio}
Online service (\url{https://convertio.co/ocr/})
\begin{itemize}
\item Fig.~\ref{Fig-5}: We uploaded two file formats. A PNG file (page
  1) and a PDF file (page 1) from the same PDF file. It was odd that
  the service produced different page segmentation and different
  accuracy results.  The OCR result from the PDF file recorded 90\%
  accuracy rate, while the OCR result from the PNG file missed 40\% of
  characters (i.e. accuracy dropped to 50\%). For both OCR results,
  the service returned acceptable page segmentations, but they were
  different. The online service time to get the file uploaded and
  converted to OCR text was about 5 minutes. (Tested April 2020)
\item Fig.~\ref{Fig-6}: It took us 10 minutes to upload and analyze
  the PDF of size 303 kb. It displayed an error message ``Analyze
  error''. (Tested March 2020)
\end{itemize}

\subsection{Testing on Arabic and Persian}
\subsubsection{Adobe Acrobat Pro DC}
\begin{itemize}
\item Does not support OCR for Arabic, Persian or Pashto languages.
\end{itemize}
  
\subsubsection{NovoVerus}
We received OCR tests from the company for three Arabic and Persian
documents. The company does not offer an evaluation version, but they
were kind enough to send us OCRed results back. Generally speaking,
the overall results were the best.

\begin{itemize}
\item Fig.~\ref{Fig-7}: Accuracy was 85\%+ (Tested Feb 2020). See Persian Sample Page with OCR text 
\item Note: Red - error, Purple - shall split; \arabicfontorimg{ک}{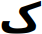} should be \arabicfontorimg{گ}{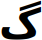}
\end{itemize}
  
\subsubsection{ABBYY}
\begin{figure}
  \includegraphics[width=0.99\textwidth]{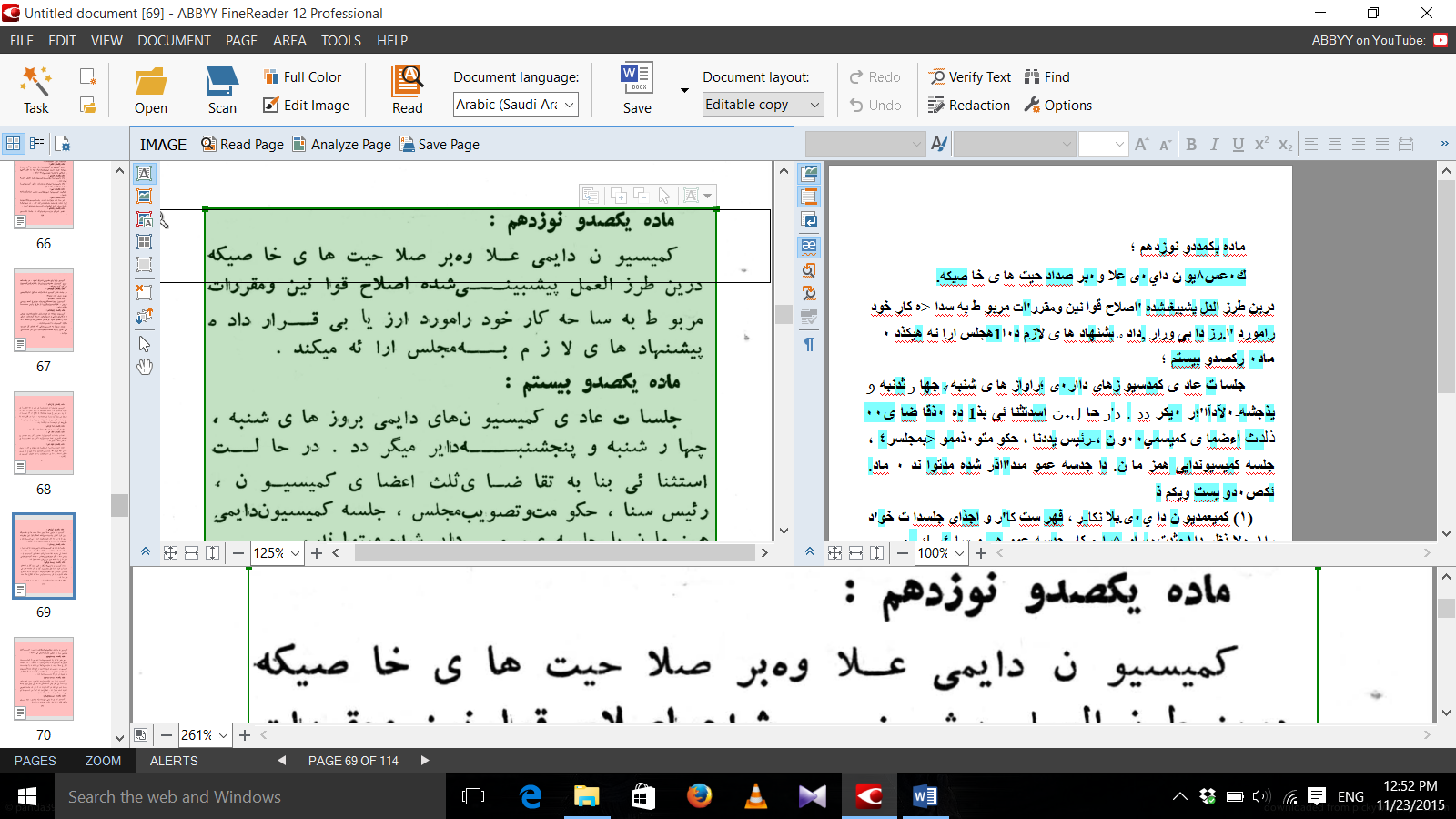}
\caption{\label{Fig-7}Original Arabic text (simple layout) and OCR result}
\end{figure}

FineReader 12 Professional: 
\begin{itemize}
\item Fig.~\ref{Fig-7}: Many characters were misread or replaced. Accuracy was not satisfactory. (Tested Nov 2015).
\end{itemize}
  
\begin{figure}
  \includegraphics[width=0.99\textwidth]{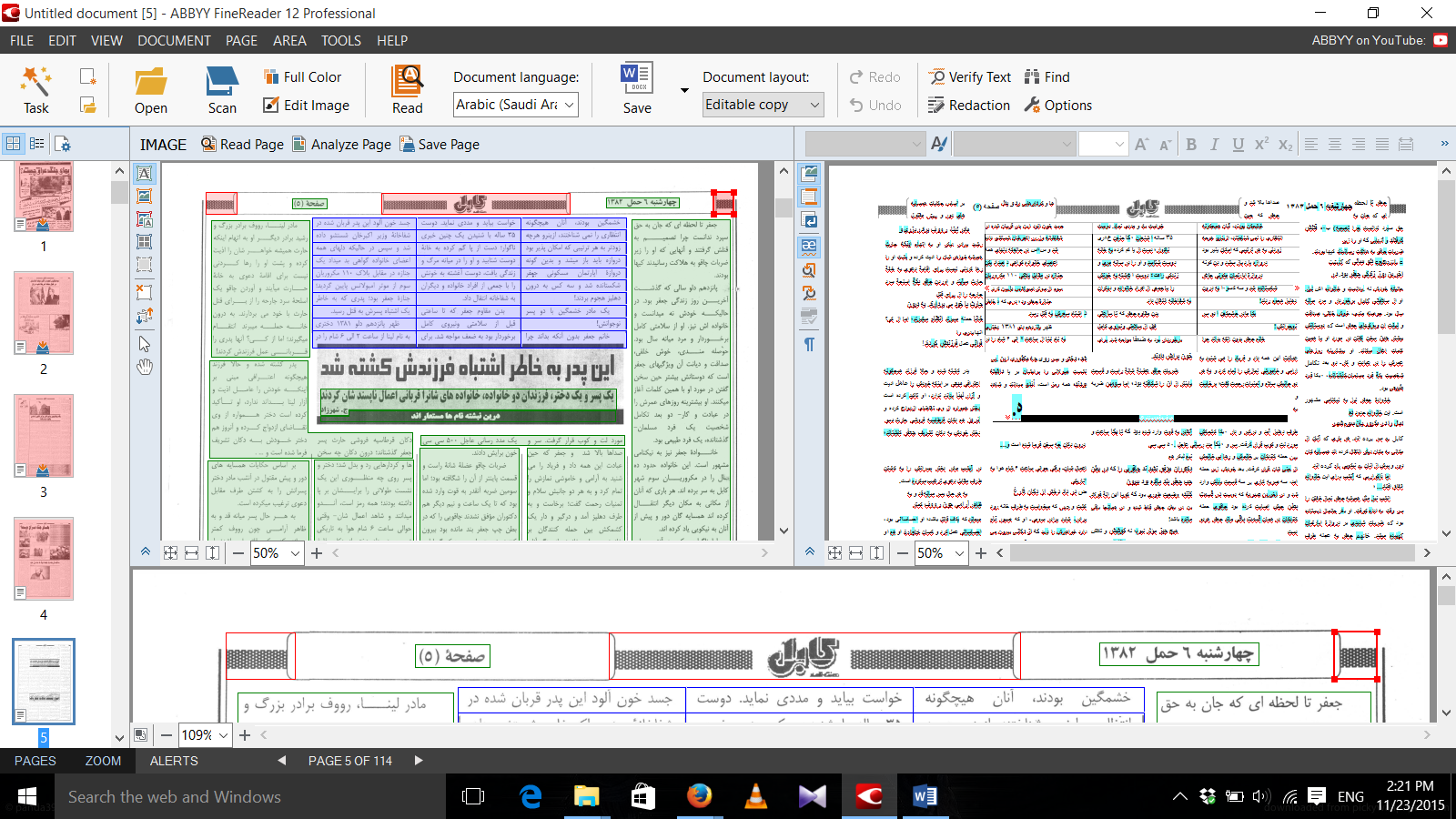}
  \caption{\label{Fig-8}: Original Arabic text (complex layout) and OCR result}
\end{figure}

\subsubsection{Kraken}
In September 2017, Mohamad Moussa of our team evaluated Kraken for
Arabic language. The sample text has 261 words and 44 of them were
wrong, meaning 83\% word accuracy and 95\% character accuracy,
because most of these errors are one character error.

\subsection{Observations on evaluated OCR systems and services}
We conducted limited tests on Chinese materials, including
contemporary materials in both Simplified Chinese characters (images
from current online newspapers) and Traditional Chinese characters
(images from current Taiwan newspapers), and retrospective materials
from the 1600s to 1940,s whose images were obtained by
digitization. The contemporary materials typically have modern writing
(left-to-right first, and then top-to-down order) with current fonts,
while retrospective materials have old writing methods (top-to-down
first, and then right-to-left) and are either on old fonts or
handwriting.

The evaluation results show that the accuracy rates for Traditional
Chinese characters vary, depending on the source of traditional Chinese
characters. We processed some samples from a contemporary Traditional
Chinese newspaper. In this case, for the current materials (simple layout
and current fonts) the accuracy rate was as high as 90-95\%.

%\subsection{Evaluations}

Based on the experiments we carried out, we can make these general conclusions:
\begin{enumerate}
\item Achieving accurate page segmentation for complex layouts is
  the first priority, as failure at this stage is predictive
  of very poor overall results; the problem is better solved
  by some commercial systems.
\item Line and character segmentation has secondary priority; most
  systems appear to divide text into lines, and for non-cursive scripts
  reliably identify.
\item Using more training data does improve accuracy; thus the performance of
  OCR systems can be improved without changing their source code, if user-trained
  models are supported (e.g., for \emph{Kraken} and \emph{Tesseract}).
\end{enumerate}

\subsubsection{Tesseract 3.0 and 4.x}
We reviewed the code and architecture and did a few tests in Oct 2017. Our evaluation of this
version shows that:
\begin{itemize}
\item The system is very big with approximately 150,000 lines of C++
  code\footnote{An automated count on recent version tesseract
    4.1.0-rc1-188-gd49c6 yields 149,482 lines of C++ code in the src
    folder and 60,490 lines in header files, for a total of 209,972
    lines of code.}.  We surmise that some lines of C++ code are not
  necessary. It seems some part of the code is for version 3 and could
  be removed for version 4. 
\item Tesseract appears to have its own implementation of LSTM in C++.
\item There is a training script ``tesstrain.sh'' in folder
  ``training''. This is a nice feature that allows users to train the
  software via their own data.
\item The production version comes with pre-trained data, and
  un-trained data having 300MB space which is significant.
\item  Overall, Tesseract seems to be more open than Kraken. 
\item  The bottleneck of all systems include:
  \begin{enumerate}
  \item \textbf{Line and page segmentation:} If this version can read the first
    character, then most likely it can read the following characters.
  \item \textbf{Fonts:} Current fonts are very likely to be recognized by
    the system.  Traditional ones or handwriting fonts are not.
  \end{enumerate}
\end{itemize}

\begin{comment}

NOTE: This is no longer needed, as it is included in the BibTeX file. (Marek)

Reference
   
Walker J., Fujii Y., Popat A. C.. 2018. "A Web-Based OCR Service for Documents". The 13th IAPR International Workshop on Document Analysis Systems. https://das2018.cvl.tuwien.ac.at/media/filer_public/85/fd/85fd4698-040f-45f4-8fcc-56d66533b82d/das2018_short_papers.pdf 

Smith, Ray. 2007. "An Overview of the Tesseract OCR Engine". The 9th International Conference on Document Analysis and Recognition (ICDAR) (https://static.googleusercontent.com/media/research.google.com/en//pubs/archive/33418.pdf)

Tafti, Ahmad et. al. 2016.. "OCR as a Service: An Experimental Evaluation of Google Docs OCR, Tesseract, ABBYY FineReader, and Transym" ISVC 2016. DOI: 10.1007/978-3-319-50835-1 66

\end{comment}

\subsubsection{Notes on Tesseract~4.11}
\yan {The test shows that OCR accuracy is better with feeding a whole page}
\marek{I made Yan's last comment into a preface to the tests below.}

We examined this version closer and observed that OCR accuracy is better with
feeding the program a whole page rather than individual lines,
words or characters. Further details of the tests conducted:
\begin{itemize}
\item Testing before training:
\item Training tesseract-ocr:
  \begin{itemize}
  \item Use Tesseract-ocr to generate a box file containing character
    strings and their positions in the image.
  \item Use jTessBoxEditor to double-check and correct data in the box
    file.
  \item Generate dictionary data and trained data according to the box
    file.
  \item Move the trained data to the working space of Tesseract-ocr.
  \end{itemize}
\item Testing after training:
  \begin{itemize}
  \item It cannot recognize the character if it not in the dictionary.
  \item For the character image that the dictionary includes, the
    accuracy is around 16\%.
  \end{itemize}      
\item Traditional Chinese characters from current traditional Chinese
  websites from Hong Kong and/or Taiwan: the accuracy rate is more
  than 90\%+.
\item Traditional Chinese characters from Siku QuanSu and Newspaper
  published in the 20$^{th}$ century (Fig.~\ref{Fig-1} and
  \ref{Fig-2}). The accuracy rate is relatively low.
\item Testing steps: 
  \begin{itemize}
  \item Picked 100 one-character, png format, black background and
    white character images from the project GitHub repository
    \cite{worldly-ocr-on-github} as testing data.
  \item Picked a one-page, pdf format image from ``Great Qing Legal
    Code''.
  \item The accuracy of Tesseract~OCR v.~3.0 is around 36\% for
    one-character, PNG format images, and around 70\% for the one-page
    PDF format image.
  \end{itemize}
\end{itemize}     

\subsection{Persian }
Since 2017, we have tested multiple OCR software systems for Persian. 
\subsubsection{Readiris}
\subsubsection{Pershyangar} Farsiocr \url{http://farsiocr.ir/}
Tesseract official trained data: (\url{https://github.com/tesseract-ocr/tessdata})
Wordly-ocr: (\url{https://github.com/mrychlik/worldly-ocr})

%%% Local Variables:
%%% mode: latex
%%% TeX-master: "main.tex"
%%% TeX-engine: luatex
%%% End:

%% file: video-catalogue.tex
\section{Videos on YouTube}
Most of the progress achieved by our project is illustrated in a
series of videos posted on YouTube. In Table~\ref{tab:video-catalogue} one
finds a list of the videos, including the Web links. Brief summaries accompany the
content, but the videos should be mostly self-explanatory. Loosely, we
will present the videos in antichronological order.

\begin{table}[htb]
  \caption{\label{tab:video-catalogue}}
  \begin{tabularx}{\textwidth}{|X|l|}
    \hline\hline
    Video Title & Link \\
    \hline\hline
    Image-to-text conversion for Farsi & \url{https://youtu.be/-TTZ7w3ujEE} \\
    \hline
    Comparing Chinese Characters Ignoring Damage & \url{https://youtu.be/O3FPrgrPbhA}\\
    \hline
    Comparing two Chinese Characters in the Presence of Damage& \url{https://youtu.be/f6aQdm4hgX0}\\
    \hline
    OCR on Latin Characters - Clustering& \url{https://youtu.be/1bKyYMoC8Zs}\\
    \hline
    Chinese OCR - Nearest Neighbors& \url{https://youtu.be/srVzYrVg4zI}\\
    \hline
    Dynamic Time Warping& \url{https://youtu.be/XcbZL4KECRQ}\\
    \hline
    Arabic/Persian OCR - baseline& \url{https://youtu.be/PGMQI-N41oo}\\
    \hline
    OCR on Rotated and Reflected Numbers& \url{https://youtu.be/ZmhUdBv4bLE}\\
    \hline
    Chinese character OCR with scale bound& \url{https://youtu.be/ft3OkJEh5fQ}\\
    \hline
    Rotated Text to Helvetica& \url{https://youtu.be/LV3bu8U2b9E}\\
    \hline
    OCR on Traditional Chinese characters& \url{https://youtu.be/hws9uVjISsw}\\
    \hline
    OCR On Rotated Text& \url{https://youtu.be/1icnxxl2cKU}\\
    \hline
    Converting Newsprint to Times New Roman& \url{https://youtu.be/TogDl_JUl-0}\\
    \hline
    Chinese OCR - clustering of characters in 7 pages& \url{https://youtu.be/Qgn0aRDvD3o}\\
    \hline
    English Clusters& \url{https://youtu.be/URzOuHpsN-g}\\
    \hline
    OCR Test Content& \url{https://youtu.be/9SHi9057F7A}\\
    \hline
    Newsprint to Times New Roman - OCR& \url{https://youtu.be/YenxuIBAU-g}\\
    \hline
    Traditional Chinese OCR - Two books& \url{https://youtu.be/-_cD2VAsyW8}\\
    \hline
    Mapping Times New Roman to Helvetica& \url{https://youtu.be/gEiHhoJ9HzU}\\
    \hline
    ChineseOCRMovie& \url{https://youtu.be/2VHX5HnZHaY}\\
    \hline
    A "Catscan" of an Artificial Neural Network& \url{https://youtu.be/tEyRGVuEgh4}\\
    \hline
    A "Catscan" of an Artificial Neural Network& \url{https://youtu.be/ej8sNN3YrAM}\\
    \hline
    OCR on a document in Persian& \url{https://youtu.be/T4sALF10gUs}\\
    \hline
    ChinesePageSegmentation& \url{https://youtu.be/eCO8IVx4lCk}\\
    \hline
    Chinese Character Search& \url{https://youtu.be/L2YxqLnzATg}\\
    \hline\hline
  \end{tabularx}
\end{table}

%%% Local Variables:
%%% mode: latex
%%% TeX-master: "main.tex"
%%% TeX-engine: luatex
%%% End: